%% file: main.tex
\documentclass[runningheads]{llncs}

\usepackage{eccv}

\usepackage{eccvabbrv}

\usepackage{graphicx}
\usepackage{booktabs}

\usepackage[accsupp]{axessibility}  

\usepackage{hyperref}

\usepackage{orcidlink}

\begin{document}

\title{ConceptWeaver: Weaving Disentangled Concepts with Flow} 

\titlerunning{ConceptWeaver}

\author{Jintao Chen\inst{1,2}$^{*}$ \and
Aiming Hao\inst{2}$^{*}$ \and
Xiaoqing Chen\inst{2} \and
Chengyu Bai\inst{1} \and
Chubin Chen\inst{2} \and
Yanxun Li\inst{2} \and
Jiahong Wu\inst{2}$^{\dagger}$$^{\ddagger}$  \and
Xiangxiang Chu\inst{2} \and
Shanghang Zhang\inst{1}$^{\ddagger}$ 
}

\authorrunning{chen et al.}


\institute{Peking University \and AMAP, Alibaba Group}


\maketitle
\begingroup
\renewcommand{\thefootnote}{}
\footnotetext{\textsuperscript{*} Equal contribution. \textsuperscript{$\dagger$} Project Lead. \textsuperscript{$\ddagger$} Corresponding authors.}
\addtocounter{footnote}{-1}
\endgroup
\vspace{-5pt}
\begin{center}
\includegraphics[width=1.0\linewidth]{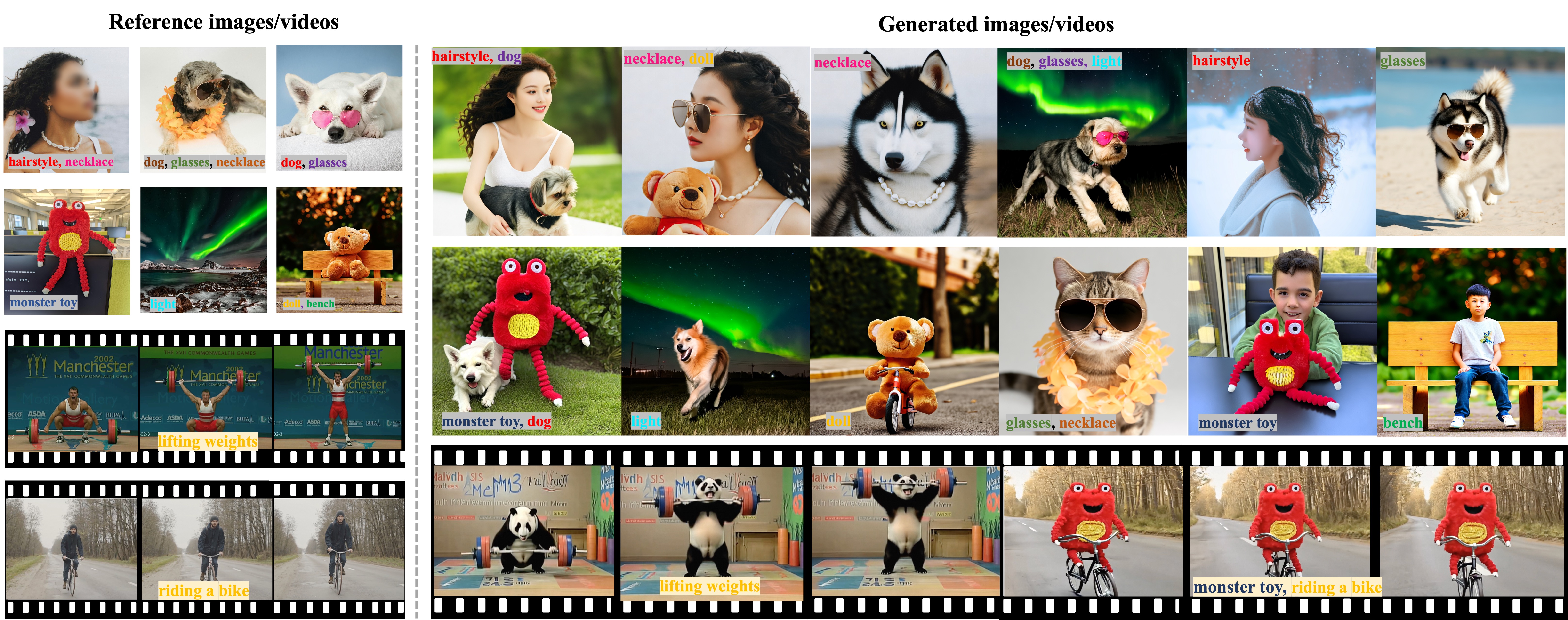}
\vspace{-25pt}
\end{center}
\captionof{figure}{\textbf{Compositional Synthesis with ConceptWeaver.} Our framework learns a visual concept, like a shirt pattern, from a single reference image by optimizing a personalized semantic offset. This offset is then injected the concept into new scenes by our stage-aware CWG, enabling compositional synthesis across diverse contexts.}
\label{fig:teaser}

\input{sections/0_abs}
\input{sections/1_intro}

\input{sections/2_related_works}

\input{sections/3_pri}

\input{sections/3_methods}
\input{sections/4_expriments}

\input{sections/5_conclusion}

%
%
\bibliographystyle{splncs04}
\begingroup
\sloppy
\emergencystretch=2em
\hbadness=1200
\bibliography{main}
\endgroup

\input{sections/X_supp}

\end{document}

%% file: sections/0_abs.tex
\begin{abstract}
\vspace{-10pt}

Pre-trained flow-based models excel at synthesizing complex scenes, yet lack a direct mechanism for disentangling and customizing their underlying concepts from one-shot real-world sources. To demystify this process, we first introduce a novel differential probing technique to isolate and analyze the influence of individual concept tokens on the velocity field over time. This investigation yields a critical insight: the generative process is not monolithic but unfolds in three distinct stages. An initial \textbf{Blueprint Stage} establishes low-frequency structure, followed by a pivotal \textbf{Instantiation Stage} where content concepts emerge with peak intensity and become naturally disentangled, creating an optimal window for manipulation. A final concept-insensitive \textbf{Refinement Stage} then synthesizes fine-grained details. Guided by this discovery, we propose \textbf{ConceptWeaver}, a framework for one-shot concept disentanglement. ConceptWeaver learns concept-specific semantic offsets from a single reference image using a stage-aware optimization strategy that aligns with the three-stage framework. These learned offsets are then deployed during inference via our novel ConceptWeaver Guidance (CWG) mechanism, which strategically injects them at the appropriate generative stage. Extensive experiments validate that ConceptWeaver enables high-fidelity, compositional synthesis and editing, demonstrating that understanding and leveraging the intrinsic, staged nature of flow models is key to unlocking precise, multi-granularity content manipulation.
\vspace{-8pt}

\keywords{One-shot Concept Disentanglement \and Stage-aware Pptimization \and ConceptWeaver Guidance}
\end{abstract}

\vspace{-15pt}

%% file: sections/1_intro.tex
\section{Introduction}
\label{sec:intro}

Modern generative models~\cite{wan2025wan, yang2024cogvideox, esser2024scaling,flux2024,hacohen2024ltx}, functioning like digital looms, can weave together diverse elements---objects, attributes, and motions---into highly realistic visual tapestries from text prompts. This remarkable capability implies an underlying compositional structure in their internal representations. However, moving beyond simply generating what is in a prompt to precisely controlling how each concept appears remains a central challenge. Achieving true personalized content creation requires the ability to disentangle these visual building blocks, extracting them from reference images and seamlessly recombining them in novel compositions. This is the key to unlocking the next level of creative and controllable synthesis~\cite{zhang2025flexiact, avrahami2025stable,tewel2023key}.

Early personalization techniques, such as textual inversion~\cite{gal2022image} or parameter fine-tuning~\cite{hu2022lora,jin2025latexblend,ruiz2023dreambooth,ling2024motionclone}, achieve single-concept embedding but often face a fundamental trade-off between subject fidelity and prompt controllability. More recent works~\cite{garibi2025tokenverse,zhong2025mod,chen2025xverse,po2024orthogonal,huang2025resolving} have sought to disentangle multiple concepts from a single image. A significant breakthrough in this area is TokenVerse~\cite{garibi2025tokenverse}, which leverages a specific architectural feature for control: the per-token modulation space---a pathway in certain Diffusion Transformer (DiT)~\cite{peebles2023scalable} variants (e.g., Flux~\cite{flux2024}) where text embeddings directly modulate channels. By learning offsets in this space, TokenVerse enables impressive, mask-free multi-concept personalization.

Despite its strengths, TokenVerse has two inherent \textbf{limitations}. First, it depends on a text-conditioned modulation path, an architectural feature absent from most DiTs~\cite{peebles2023scalable}, where modulation typically encodes only non-semantic signals such as the timestep $t$. This reliance constrains its applicability to a narrow subset of models. Second, by focusing solely on a spatial locus of control (i.e., which parameters to modify), it neglects the temporal dimension of concept formation---failing to address when a concept should be manipulated to maximize effect and minimize interference. This dual gap in architectural generality and temporal awareness underpins the motivation for our work.

Building on these observations, our work addresses both the temporal challenge and the generality gap identified above. We begin with a novel differential probing technique to isolate and chart the influence of individual concepts on the velocity field~\cite{lipman2022flow,liu2022flow} over time. This analysis yields a critical and universal insight: \emph{the generative process is not monolithic but unfolds in a distinct three-stage framework}. An initial \textbf{Blueprint Stage} establishes coarse structure, followed by a pivotal \textbf{Instantiation Stage} where content concepts emerge with peak intensity and become naturally disentangled. A final, concept-insensitive \textbf{Refinement Stage} then adds fine-grained detail. This discovery reveals a principled, optimal window for manipulation.

Guided by this temporal insight, we introduce \textbf{ConceptWeaver}, a one-shot learning and guidance framework that operationalizes the probing results without requiring access to a modulation space. ConceptWeaver learns concept-specific semantic offsets from a single reference data using a stage-aware optimization strategy that concentrates training exclusively on each concept’s peak-influence stage. During inference, our ConceptWeaver Guidance (CWG) mechanism injects these learned offsets only during the corresponding stage, aligning control with the model’s intrinsic dynamics rather than imposing uniform, heuristic constraints. By combining architecture-agnostic applicability with precise temporal alignment, ConceptWeaver enables high-fidelity \mbox{compositional synthesis.}

In summary, our contributions are as follows:
\begin{itemize}
    \item  A novel differential probing technique that reveals, for the first time, a universal three-stage framework for concept formation in flow-based models, identifying a principled window for manipulation.
    \item ConceptWeaver, a one-shot framework featuring a stage-aware optimization strategy and a novel CWG mechanism, designed to leverage this temporal discovery for principled concept control.
    \item Extensive experiments demonstrating superior performance in high-fidelity, compositional synthesis, proving the efficacy of our approach.
\end{itemize}

%% file: sections/2_related_works.tex
\section{Related Works}
\label{sec:related}

\paragraph{\textbf{Flow Methods.}} Diffusion models~\cite{ho2020denoising, song2020denoising, song2020score, mao2025omni} have revolutionized generative tasks with high-quality outputs, but their inverse processes rely on Markov chains leading to slow sampling, and score-based formulations require expensive gradient estimations for high-dimensional data. To address these limitations, Flow Matching (FM) was proposed as a lightweight alternative framework ~\cite{flux2024, lipman2024flow,lipman2022flow,gat2024discrete,liu2022flow}. It bypasses explicit noise addition steps and directly learns a continuous flow mapping noise distributions to data distributions---optimizing a velocity field that guides smooth transitions from random noise to real data. Unlike diffusion models, FM enables efficient sampling via ODE solvers (faster than Markov chains) and avoids gradient-based score estimation. Subsequent works~\cite{fan2025cfg,li2025flowdirector,xie2025dnaedit, bai2025uniedit} extended FM to conditional tasks: FlowEdit~\cite{kulikov2025flowedit} leverages conditional Flow Matching to enable precise image editing by aligning source and target velocity fields. It uses the difference between source and target velocity fields to guide inverse ODE sampling\cite{lu2022dpm}, solving the control issues in diffusion-based editing (e.g., drift from desired edits).

\vspace{-10pt}

\paragraph{\textbf{Personalization Methods.}} Personalized generation~\cite{gal2022image,hu2022lora,jin2025latexblend,ruiz2023dreambooth,ling2024motionclone,gu2023mix, liu2025phantom} aims to synthesize images that faithfully incorporate a user-specified concept (e.g., a person, object or style) while adhering to a given textual description \cite{jiang2025vace, wang2025dualreal,zhang2025flexiact, wei2024dreamvideo, mao2025omni}. This task typically relies on a small set of reference images to generalize the target concept into new contexts. Existing personalization methods are generally categorized by three inversion spaces. In the \textit{textual space}, the concept is represented as a learnable pseudo-word embedding (e.g., Textual Inversion~\cite{gal2022image, kumari2023multi}). The \textit{parameter space} methods fine-tune the diffusion model itself to internalize the concept, with DreamBooth~\cite{ruiz2023dreambooth} being a prime example, often augmented by parameter-efficient techniques like LoRA~\cite{hu2022lora}. Despite these advancements, a core challenge remains: balancing subject fidelity with text controllability. Many methods that excel at capturing fine-grained visual details often overfit, leading to outputs that merely reproduce original images rather than adapting to prompt-specified structural or contextual variations---a limitation known as the ``copy-paste phenomena''~\cite{liu2025phantom, chen2025multi,polyak2024movie}.

\vspace{-10pt}


\paragraph{\textbf{Concept Decoupling.}} Recent personalized generation efforts aim to extract and reuse multiple visual concepts from minimal reference data~\cite{garibi2025tokenverse,jin2025latexblend, chen2025xverse, zhong2025mod,chen2025taming}. However, cleanly separating semantically distinct attributes (e.g., shape, appearance, pose, context) remains challenging. For example, Break-a-Scene~\cite{avrahami2023break} uses manually annotated spatial masks for multi-concept learning, which limits scalability and conflates non-spatial attributes not addressable by masks. Alternative approaches like Inspiration Tree~\cite{vinker2023concept} and ConceptExpress~\cite{hao2024conceptexpress} associate a single image with several learned tokens. Yet, the implicit semantic assignment to individual tokens during optimization often lacks user control, leading to unclear or inconsistent visual facets governed by each token, which hinders reliable composition.

%% file: sections/3_pri.tex
\section{Preliminaries and Analysis}

\subsection{Conditional Velocity Field}
Flow models~\cite{wan2025wan,flux2024,lipman2022flow,liu2022flow} generation as learning a continuous-time velocity field that transports samples from a simple prior distribution to the target data distribution. Given a data example $\boldsymbol{x}_0\sim \mathcal{D}$ and its fully noised counterpart $\boldsymbol{x}_1\sim \mathcal{N}\left(0,\boldsymbol{I}\right)$, the forward trajectory is defined by:
\begin{equation}
\boldsymbol{x}_t=\left(1-t\right)\boldsymbol{x}_0+t\boldsymbol{x}_1,\ t\in \left[0,1\right],
\end{equation}
yielding the velocity:
\begin{equation}
\boldsymbol{v}_t=\frac{\partial \boldsymbol{x}_t}{\partial t}=\boldsymbol{x}_1-\boldsymbol{x}_0.
\end{equation}
In the text‑conditioned setting, let $\boldsymbol{c}$ denote the text prompt. The prompt is processed by a text encoder into a sequence of token embeddings $\left\{\boldsymbol{e}_1,\cdots,\boldsymbol{e}_K\right\}$, each corresponding to an individual lexical unit within $\boldsymbol{c}$. These embeddings are injected into the velocity predictor:
\begin{equation}
\hat{\boldsymbol{v}}_t=\boldsymbol{v}_{\theta}\left(\boldsymbol{x}_t,t,\left\{\boldsymbol{e}_k\right\}_{k=1}^K\right),
\end{equation}
directly modulating the instantaneous direction and magnitude of motion in latent space. This conditioning shapes the velocity field $\boldsymbol{v}_\theta(\cdot)$ into a semantic trajectory aligned with the prompt, but also intrinsically \emph{couples} all concepts in $\boldsymbol{c}$ into a single, entangled guidance force.

\begin{figure*}[!t]
  \centering
    \includegraphics[width=0.95\linewidth]{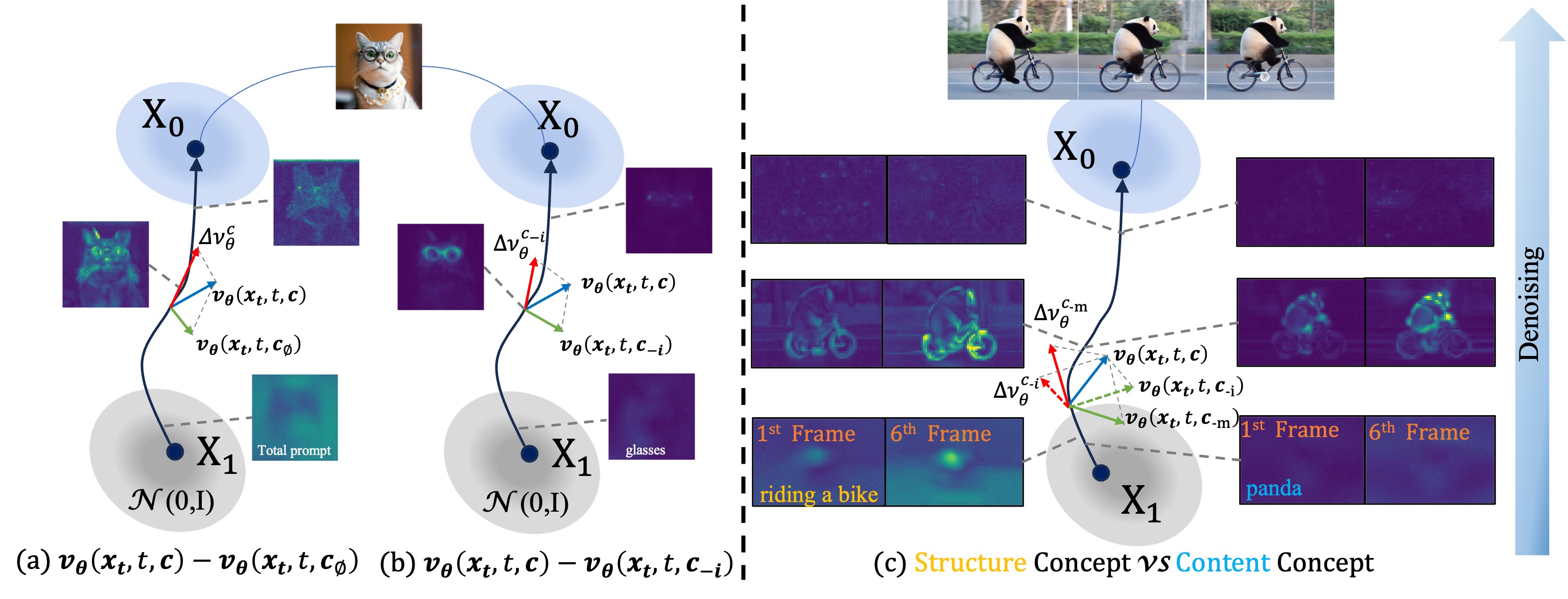}
    \vspace{-10pt}
    \caption{\textbf{Probing Concept Formation Dynamics.} While standard prompt guidance (a) provides a dense, entangled signal, our differential probing technique (b) isolates the dynamic influence of individual concepts. This analysis uncovers a consistent three-stage framework (c): an early \textbf{Blueprint Stage} dominated by structural concepts (e.g.,``riding a bike''), a pivotal mid-stage \textbf{Instantiation Stage} where content concepts (e.g., ``panda'') are decoupled and peak in intensity, and a final \textbf{Refinement Stage} where signals fade.}
    \label{fig:velshift}
    \vspace{-10pt}
\end{figure*}

\subsection{Prompt Semantic Guidance}
When stronger prompt adherence is desired, one can operate directly in the velocity field to reinforce the semantic trajectory induced by the text. Among such approaches, Classifier-Free Guidance (CFG)~\cite{ho2022classifier,chen2025s} is widely adopted. CFG computes a single \emph{global semantic shift} in the velocity field by contrasting the model’s velocity predictions guided by the text prompt with those generated from a fully null prompt, and then amplifies this shift during inference:
\begin{equation}
\boldsymbol{v}_{\text{CFG}} = \boldsymbol{v}_{\theta}(\boldsymbol{x}_t, t, \boldsymbol{c}_{\emptyset}) + w \cdot \left(\boldsymbol{v}_{\theta}(\boldsymbol{x}_t, t, \boldsymbol{c}) - \boldsymbol{v}_{\theta}(\boldsymbol{x}_t, t, \boldsymbol{c}_{\emptyset})\right),
\end{equation}
where $w>1$ controls the guidance scale. The difference term $\Delta \boldsymbol{v}_{\text{prompt}} = \boldsymbol{v}_{\theta}(\boldsymbol{x}_t, t, \boldsymbol{c}) - \boldsymbol{v}_{\theta}(\boldsymbol{x}_t, t, \boldsymbol{c}_{\emptyset})$ acts as a prompt semantic guidance in velocity field, steering the generative trajectory toward the overall semantics of $\boldsymbol{c}$. 
By construction, this guidance reflects the combined influence of all concepts described in the prompt, providing holistic guidance without distinguishing their individual contributions. The inability to isolate concept-specific influence motivates the following concept disentanglement task.

\subsection{Concept Decoupling}

Real-world visual scenes are inherently compositional, blending high-level structural concepts (e.g., spatial layouts) with content concepts (e.g., identities, textures). Pre-trained flow-based generative models, trained on such data, demonstrate a remarkable ability to synthesize novel, coherent scenes by recombining these elements.

This powerful compositional capability strongly suggests that the model’s internal representations must go beyond memorizing concept combinations; coherent compositional generalization plausibly requires that constituent concepts be encoded in a partially disentangled form within the underlying velocity field. We therefore hypothesize that \emph{such a disentangled structure exists, enabling the manipulation and recomposition of individual concepts.}

The \textbf{Concept Decoupling} task examines this hypothesis in the challenging \textbf{one-shot} setting. The practical goal is to achieve fine-grained control over composition. Formally, given $N$ reference images $\left(\boldsymbol{I}_{1},...,\boldsymbol{I}_{N}\right)$, each paired with a textual prompt comprising a multiple concepts set $\mathcal{C}_i$, the task is to synthesize a novel target image $\tilde{\boldsymbol{I}}$ by selectively recombining concepts from these sets. For instance, given $\boldsymbol{I}_1$ describing “cat” and “glasses” and 
$\boldsymbol{I}_2$ describing “dog” and “a beach,” the goal is to generate $\tilde{I}$ depicting “dog” (from $\mathcal{C}_2$) wearing “glasses” (from $\mathcal{C}_1$).

Achieving this level of control, however, requires a deeper understanding of the model's internal dynamics. This reframes the challenge: the central analytical task becomes to characterize the individual contribution of each semantic token to the conditional velocity field. A significant barrier here is the model's inherent non-linearity; the full velocity field likely involves complex interactions, making a simple additive decomposition insufficient~\cite{hertz2022prompt,liang2024diffusion}. This exposes a fundamental question at the heart of generative control: \emph{How can we reliably isolate and quantify the contribution of an individual concept, when its influence is deeply entangled with others in a dynamic, non-linear system?} Answering this question would allow us to chart the temporal lifecycle of concepts---their emergence, peak contribution, and decay---providing the necessary foundation for principled, fine-grained compositional control.

%% file: sections/3_methods.tex
\section{Methods}
\label{sec:methods}

Our method for one-shot concept disentanglement proceeds in three steps. First, we introduce a probing technique to analyze the temporal dynamics of concept formation (Sec.~\ref{sec:probing}). Guided by this analysis, we then present \textbf{ConceptWeaver}, a framework for learning concept-specific offsets from a single reference (Sec.~\ref{sec:conceptweaver}). Finally, we detail our stage-aware guidance mechanism that applies these offsets for high-fidelity compositional synthesis (Sec.~\ref{sec:cwg}).

\begin{figure*}[t]
  \centering
  \includegraphics[width=0.90\linewidth]{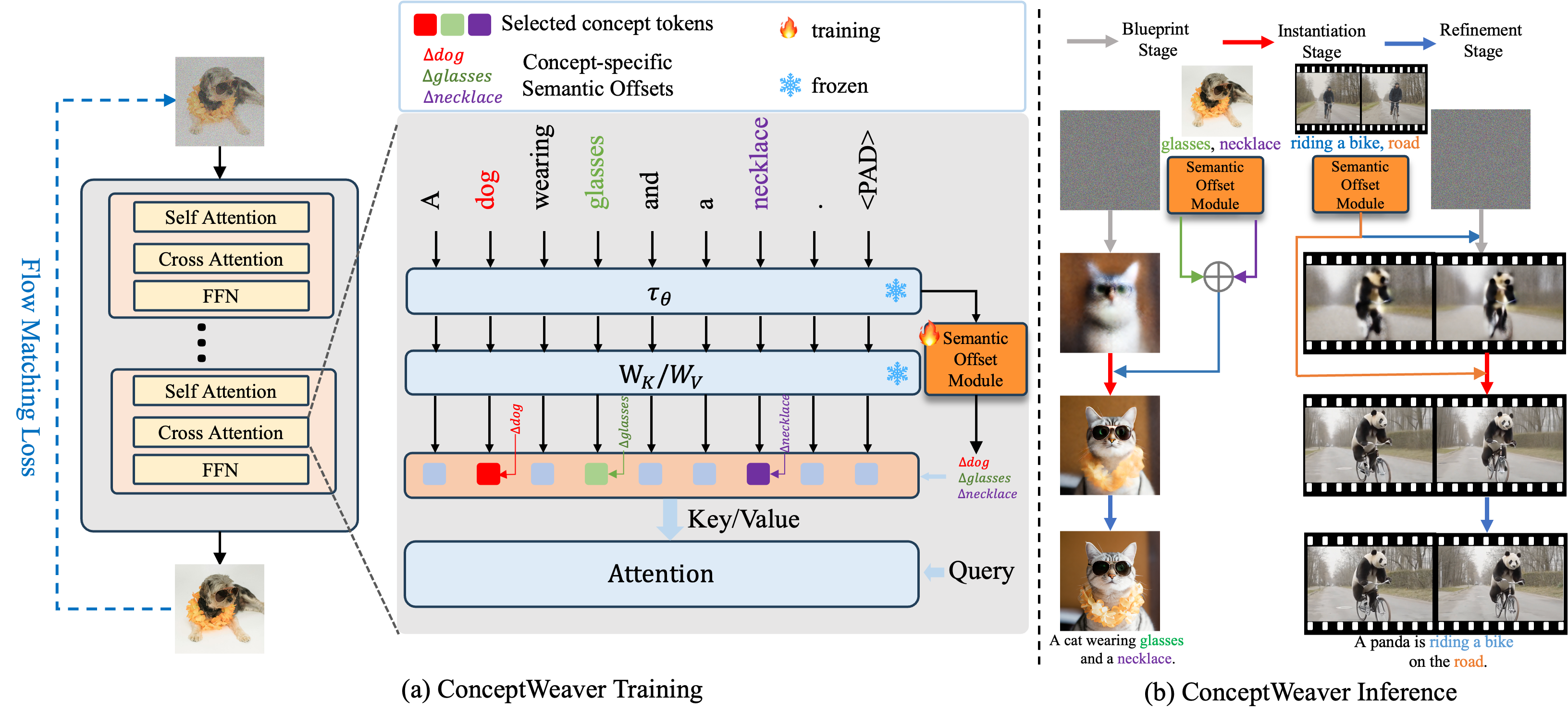}
  \vspace{-10pt}
  \caption{\textbf{Flowchart of proposed ConceptWeaver.} Our framework consists of two main phases: training and inference. (a) \textbf{During training}, a lightweight Semantic Offset Module learns to represent a visual concept by adding a learnable offset to the key/value pairs of its corresponding concept token. This is optimized with a stage-aware loss. (b) \textbf{During inference}, our ConceptWeaver Guidance (CWG) applies these offsets in a stage-aware manner: structural offsets are injected during the Blueprint Stage, and content offsets during the Instantiation Stage, enabling precise compositional control.}
  \label{fig:overview}
  \vspace{-18pt}
\end{figure*}

\subsection{Concept-Specific Velocity Field Probing}
\label{sec:probing}
To investigate how individual concept $\boldsymbol{e}_i$ contributes to the velocity field, we introduce a \textbf{differential probing technique} inspired by CFG. Instead of contrasting with a fully null prompt, we compute the difference against a \emph{partially-masked prompt} $\boldsymbol{c}_{-i}=\boldsymbol{c}-\Delta \boldsymbol{c}_{\mathrm{concept}_i}$, where $\Delta \boldsymbol{c}_{\mathrm{concept}_i}=\boldsymbol{e}_i\cdot \boldsymbol{\delta}_i$. (only the $i$-th concept token $\boldsymbol{e}_i$ is retained and all other components are zero) This yields an estimate of the \emph{concept-specific velocity shift}:
\begin{equation}
\Delta\boldsymbol{v}_{\mathrm{concept}_i} = \boldsymbol{v}_\theta(\boldsymbol{x}_t, t, \boldsymbol{c}) - \boldsymbol{v}_\theta(\boldsymbol{x}_t, t, \boldsymbol{c}_{-i}).
\end{equation}
which serves as an approximate measure of the dynamic influence of $\boldsymbol{e}_i$ throughout generation, despite the non-linearity of the underlying model.

For our analysis of both image and video generation, we utilize \textbf{Wan2.2}~\cite{wan2025wan} as our base Flow Matching model. As shown in Figure~\ref{fig:velshift}, such probing reveals temporal variations that are hidden under standard holistic guidance. While the standard shift $\Delta\boldsymbol{v}_{\mathrm{prompt}}$ in (a) provides a dense, entangled signal throughout generation, the concept-specific shift $\Delta\boldsymbol{v}_{\mathrm{glasses}}$ in (b) evolves dynamically: starting as a weak, diffuse signal in early timesteps, peaking sharply and being tightly localized at mid-stage, and then fading into sparse high-frequency noise in later timesteps. Beyond this example, our probing shows a clear distinction between structural and content concepts. Structural concepts (e.g., “riding a bike”) exert their strongest influence in the early-to-mid stages, establishing the scene’s layout and motion. Content concepts (e.g., “panda” or “glasses”) follow a more concentrated lifecycle — weak and diffuse at first, sharply peaking in both intensity and spatial localization at mid-stage, then rapidly diminishing during refinement.

This directly observed, non-monotonic behavior allows us to move beyond a simple ``coarse-to-fine'' understanding and characterize the generative process through a more structured, three-stage framework:
\begin{itemize}
\item \textbf{Blueprint Stage} (early $t$): Concepts exhibit weak, diffuse signals across the image as the model focuses on establishing a coarse structural layout, providing a foundation for subsequent concept refinement.
\item \textbf{Instantiation Stage} (mid $t$): The content concept reaches peak intensity and strong spatial localization, marking a critical window where its semantic identity is fully materialized. In this stage, the signal is potent and disentangled from other elements, allowing precise and independent manipulation without affecting unrelated regions.
\item \textbf{Refinement Stage} (late $t$): Concept signals gradually diminish while the model concentrates on synthesizing fine-grained textures and enhancing global coherence.
\end{itemize}
This discovery---that \textbf{concepts are not uniformly exerted but rather emerge and disentangle within specific temporal stages}---forms the cornerstone of our work.

\subsection{ConceptWeaver: Learning Concept Offsets}
\label{sec:conceptweaver}
The core finding of our probing analysis is that a concept's influence can be isolated by measuring the velocity change after removing its corresponding token. This operation---effectively applying a negative offset to the prompt---motivates a new, unified perspective on concept manipulation: the principle of \textbf{concept offsets}. In our probing, the ``existence shift'' ($\text{none}\rightarrow \text{generic}$) is measured by applying a negative semantic offset $\Delta \boldsymbol{c}_{\mathrm{concept}_i}$. We extend this offset-based formulation from \emph{analysis} to \emph{synthesis}: instead of a predefined removal offset, we learn a reference-conditioned offset, $\Delta \boldsymbol{c}_{\mathrm{ref}}$, that induces a \emph{customization shift} ($\text{generic}\rightarrow \text{specific}$). To this end, we introduce ConceptWeaver, a one-shot framework to learn a set of \emph{concept-specific semantic offsets} from a single reference. As illustrated in Figure~\ref{fig:overview} (a), this is achieved with a lightweight, learnable \textbf{Semantic Offset Module}. This module learns an additive offset $\Delta \boldsymbol{e}_i$ for each selected concept token $\boldsymbol{e}_i$, which is then injected into the key/value representations of the model's cross-attention layers. These offsets, forming the semantic offsets $\Delta \boldsymbol{c}_{\mathrm{ref}}$, are optimized by minimizing a flow matching loss between the model's prediction and the ground-truth velocity, aiming to collapse the conditional velocity field toward the reference distribution:
\begin{equation}
\mathcal{L}_{flow}=\mathbb{E}_{}\left[\left\|\boldsymbol{v}_{\theta}\left(\boldsymbol{x}_t,t,\boldsymbol{c}+\Delta \boldsymbol{c}_{\mathrm{ref}}\right)-\boldsymbol{v}_t\right\|^2\right].
\end{equation}

Crucially, this optimization is \textbf{stage-aware}, directly leveraging our findings from the probing analysis. For content concepts like object identity (e.g., ``glasses'', ``necklace''), the optimization prioritizes loss computations within the pivotal Instantiation Stage. In contrast, for structural concepts like motion or layout (e.g., ``riding a bike''), it prioritizes the Blueprint Stage. By aligning the optimization with the model's own internal focus during its stage of peak influence, this targeted approach naturally forces the shift to encode a disentangled representation of the concept, without requiring an explicit disentanglement objective.

Upon convergence, this process yields a set of disentangled concept offsets, each encoding the visual attributes of a concept from the reference. These learned offsets can be deployed for novel synthesis as shown in Figure~\ref{fig:overview} (b).

\subsection{ConceptWeaver Guidance}
\label{sec:cwg}

Building upon standard CFG, we introduce ConceptWeaver Guidance (CWG), a flexible mechanism that extends the velocity field with our learned concept offsets $\left\{\Delta \boldsymbol{e}_i\right\}$. CWG refines coarse, prompt-level control into a fine-grained, concept-aware process by decomposing the guidance into a weighted sum of individual concept shifts.

Let $N$ be the number of customized concepts, each with a learned offset $\Delta \boldsymbol{c}_{\mathrm{ref},i}=\Delta \boldsymbol{e}_i\cdot\boldsymbol{\delta}_i$ and a corresponding guidance scale $w_{c,i}$. The full guidance formulation is:
\begin{equation}
\label{eq:cwg_multi}
\boldsymbol{v}_{\text{CWG}} = \boldsymbol{v}_{\theta}(\boldsymbol{x}_t, t, \boldsymbol{c}_{\emptyset})  + w \cdot \Delta \boldsymbol{v}_{\mathrm{prompt}}  + \sum_{i=1}^{N} w_{\mathrm{c},i} \cdot \Delta \boldsymbol{v}_{\mathrm{ref},i},
\end{equation}
Here, $\Delta \boldsymbol{v}_{\mathrm{prompt}}$ is the standard prompt-level shift, while $\Delta\boldsymbol{v}_{\mathrm{ref},i} = \boldsymbol{v}_{\theta}(\boldsymbol{x}_t, t, \boldsymbol{c}+\Delta \boldsymbol{c}_{\mathrm{ref},i}) - \boldsymbol{v}_{\theta}(\boldsymbol{x}_t, t, \boldsymbol{c})$ is the reference-concept-specific shift for the $i$-th concept. 

This per-concept weighting scheme provides fine-grained control over the contribution of each concept. By adjusting the guidance scales $\{w_{c,i}\}$ independently, the relative influence of each learned offset can be precisely modulated. This capability is crucial for balancing potentially competing concepts and achieving coherent compositional synthesis, particularly in multi-concept scenarios.

\begin{figure}[t]
    \centering
    \includegraphics[width=0.8\linewidth]{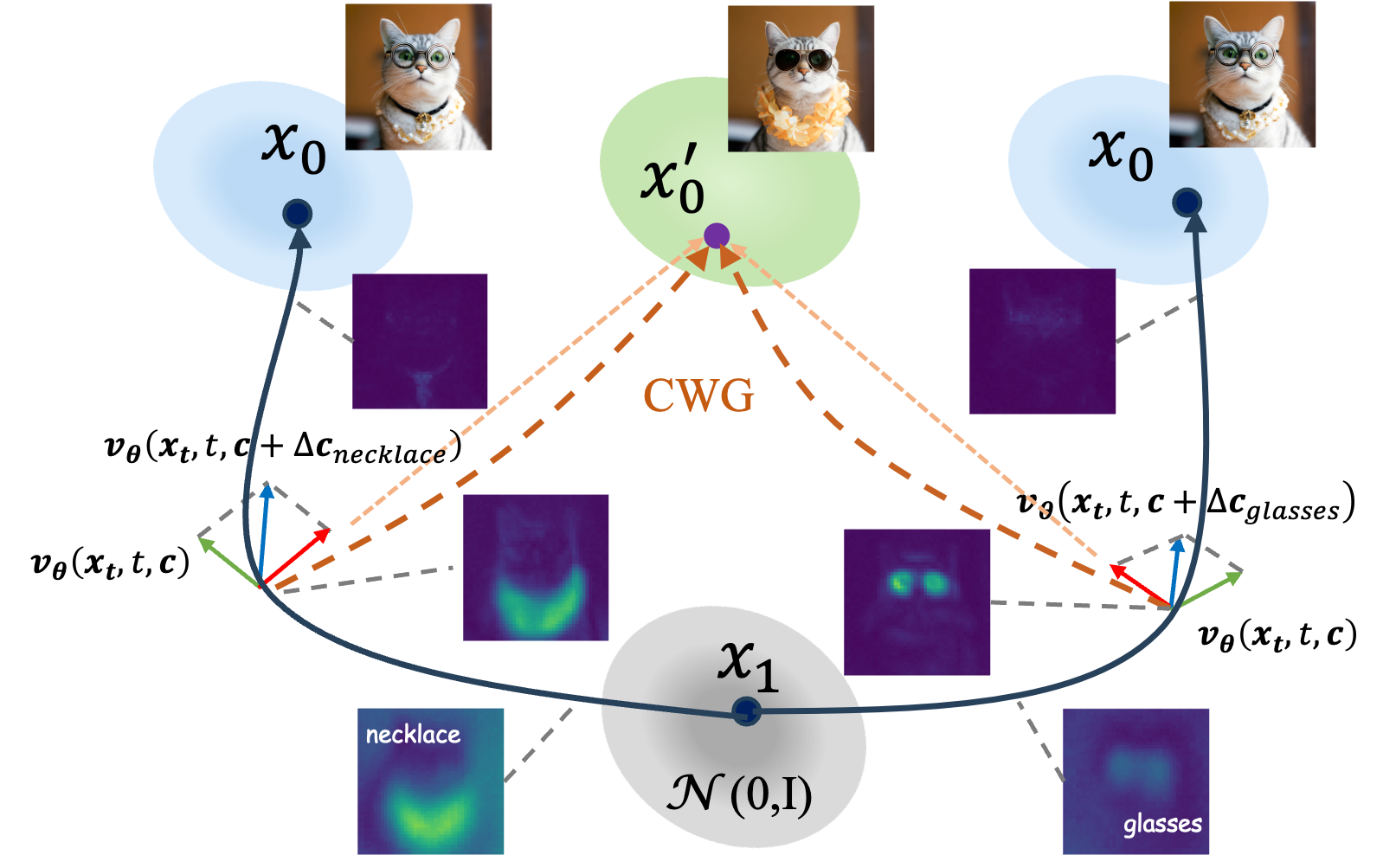}
    \vspace{-5pt}	
    \caption{\textbf{Mechanism of ConceptWeaver Guidance (CWG).} Our guidance modifies the standard generative path (to $\boldsymbol{x}_0$) by injecting learned concept shifts (red vectors) at stage-aware intervals. This timed intervention, visualized by the heatmaps, steers the trajectory from noise ($\boldsymbol{x}_1$) to a customized target ($\boldsymbol{x}_0'$), enabling precise compositional control.}
    \label{fig:cwg}
    \vspace{-20pt}	
\end{figure}

Crucially, guided by our analysis, CWG applies these learned shifts in a time-dependent manner. Structural concept offsets are injected during the \textbf{Blueprint Stage} to shape low-frequency layouts, while content concept offsets are introduced during the \textbf{Instantiation Stage} to leverage their peak signal intensity. As demonstrated in Figure~\ref{fig:cwg}, this stage-aware, multi-level guidance enables customized rendering of target concepts, faithfully preserving and integrating their attributes across diverse subjects and scenes.

%% file: sections/4_expriments.tex
\section{Expriments}

\subsection{Experimental Setups}
\paragraph{Datasets.}
For the single-concept dataset, we extracted 20 images from the Dreambench++\cite{peng2024dreambench++} and Dreambooth~\cite{ruiz2023dreambooth} datasets, and collected 10 types of motion datasets from the internet. For the multi-concept image dataset, we collected $20$ images from the internet. All our self-collected data were captioned using LLM. We used LLM to create $5$ test prompts each for single-concept generation and multi-concept fusion (including motion concept transfer).

\vspace{-10pt}

\paragraph{Evaluation Metrics.}
Consistent with prior work, our evaluation comprises Following previous studies, we evaluate our method on the following three aspects:
\begin{enumerate}
    \item \textbf{Textual Alignment:} Assessed by the average CLIP similarity ($S_{\text{CLIP}}^\mathrm{T}$)~\cite{radford2021learning} between the generated images and their textual prompts.
    
    \item \textbf{Concept Alignment:} Measured by the CLIP ($S_{\text{CLIP}}^\mathrm{I}$) and DINO ($S_{\text{DINO}}$)~\cite{zhang2022dino} similarities between the generated images and the reference images. For multi-subject generations, we report the mean similarity, calculated across individual subject-reference pairs.
    
    \item \textbf{MLLM-based Evaluation:} We employ GPT-4o\cite{hurst2024gpt} to score the alignment($ S_{GPT} $), capitalizing on its advanced visual understanding capabilities. The model is prompted to assign a score from 0 to 4 for each generated instance, based on the provided concept name(s), reference image(s), and the generated image(s). Higher scores indicate better alignment.
\end{enumerate}
\vspace{-10pt}

\paragraph{Baselines} 
Previous methods mostly focused on only one aspect. For concept decoupling and multi-concept generation in images, we choose MS-Diffusion\cite{wang2024ms} and reproduced TokenVerse\cite{garibi2025tokenverse} (as $\text{TokenVerse}^*$) and DreamBooth\cite{ruiz2023dreambooth} on Wan2.2-TI2V-5B. For concept video generation, we chose VACE-14B\cite{jiang2025vace} as a comparative method. See the Appendix \textbf{A} for specific settings.
\vspace{-10pt}

\paragraph{Implementation Details.}
In ConceptWeaver, we employ Wan2.2-TI2V-5B as our base architecture. For the images, we maintain the training resolution at $704\times 704$, inference resolution at $704\times 704$ for small number of concepts and $704\times 1280$ for multi concepts generation. We perform data augmentation methods such as random cropping and mirror flipping. For the concept videos generation, we maintained a training resolution of $704\times 1280$, $49$ frames per second (fps) at $16$. All case prompts can be expanded and enhanced using templates. The semantic offset module's rank was set to $16$, and rank alpha to $32$. The training learning rate was $1e-4$, and each case was trained for $2000$ steps.

\begin{table}[ht]
\centering
\caption{\textbf{Quantitative comparison on single-concept generation.} The metrics evaluate textual alignment ($S_{\text{CLIP}}^\mathrm{T}$) and concept alignment ($S_{\text{CLIP}}^\mathrm{I}$, $S_{\text{DINO}}$, $S_{GPT-4o}$). $\uparrow$ indicates that higher is better. The best results are highlighted in \textbf{bold}.}
\label{tab:single_concept_simple}
\begin{tabular}{l c c c c}
\toprule
Method & $S_{\text{CLIP}}^\mathrm{T} \uparrow$ & $S_{\text{CLIP}}^\mathrm{I} \uparrow$ & $S_{\text{DINO}} \uparrow$ & $S_{GPT}$ $\uparrow$ \\
\midrule 
$\text{TokenVerse}^*$ & 35.20 & 61.75 & 26.31 & 3.2 \\
MS-Diffusion & 35.02 & 75.58 & \textbf{70.69} & 3.6 \\
DreamBooth & 34.50 & 73.14 & 68.96 & 3.7 \\  
\textbf{Ours} & \textbf{35.22} & \textbf{76.13} & 69.57 & \textbf{3.8} \\
\bottomrule
\end{tabular}
\vspace{-10pt}	
\label{table:main}
\end{table}

\subsection{Main Results}
\begin{figure}[t]
    \centering
    \includegraphics[width=1.0\linewidth]{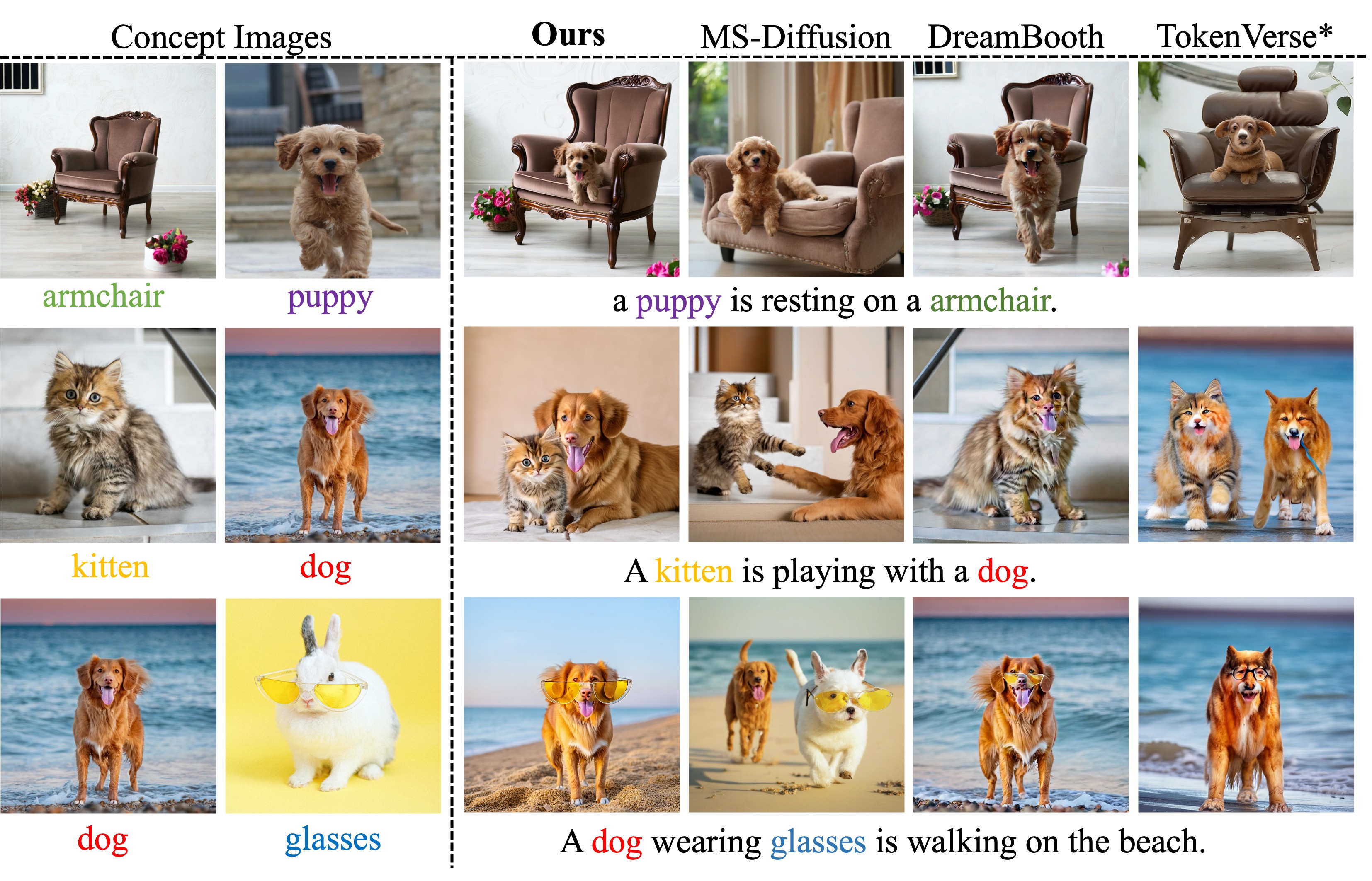}
    \vspace{-15pt}	
    \caption{\textbf{Comparison on multi-concept composition generation.} The figure illustrates different methods' performance in combining multiple concepts, focusing on overfitting to original reference concepts and the ability to disentangle target concepts.}
    \vspace{-10pt}	
    \label{fig:multi_concept_composition}
\end{figure}

\begin{figure}[ht]
    \centering
    \includegraphics[width=1.0\linewidth]{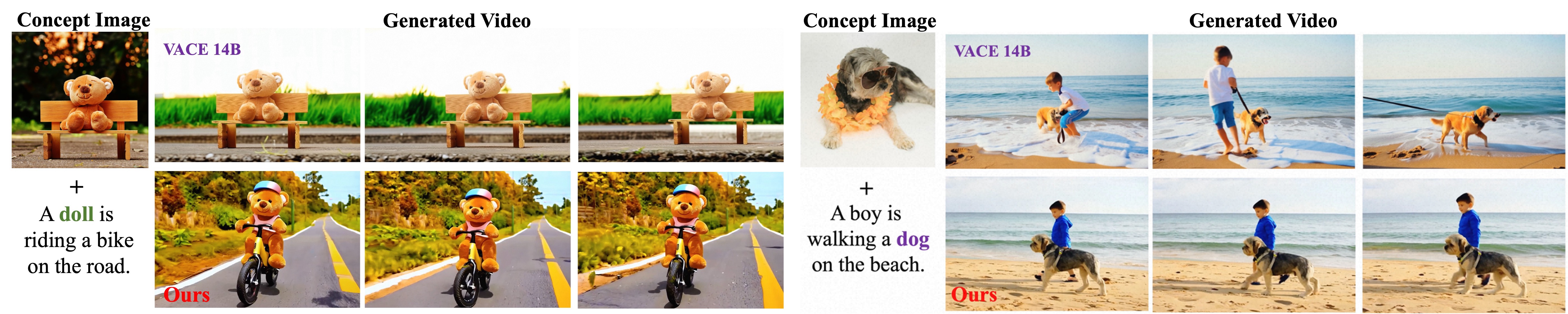}

    \caption{\textbf{Comparison of R2V task results.} Contrasts VACE method with our approach in terms of concept disentanglement, structural preservation, and ``copy-paste phenomena''.}
    \label{fig:r2v}
    \vspace{-15pt}	
\end{figure}

Our proposed method demonstrates outstanding performance across multiple generation tasks. For \textbf{single-concept generation}, as shown in Table~\ref{table:main}, our method achieves state-of-the-art results, leading in both concept alignment (S\textsubscript{CLIP}, S\textsubscript{DINO}, S\textsubscript{GPT}) and text alignment ($S_{\text{CLIP}}^\mathrm{T}$), which validates its strong fundamental generation capabilities. For \textbf{multi-concept composition}, Figure~\ref{fig:multi_concept_composition} shows that our method not only effectively integrates multiple concepts but also excels at avoiding overfitting to the reference images. Furthermore, in the \textbf{Reference To Video (R2V) task} (Figure~\ref{fig:r2v}), ConceptWeaver successfully disentangles target concepts and transfers them to new contexts. More comparison results are shown in Appendix \textbf{B}.
\vspace{-15pt}	

\begin{figure*}[t]
    \centering
    \includegraphics[width=1.0\linewidth]{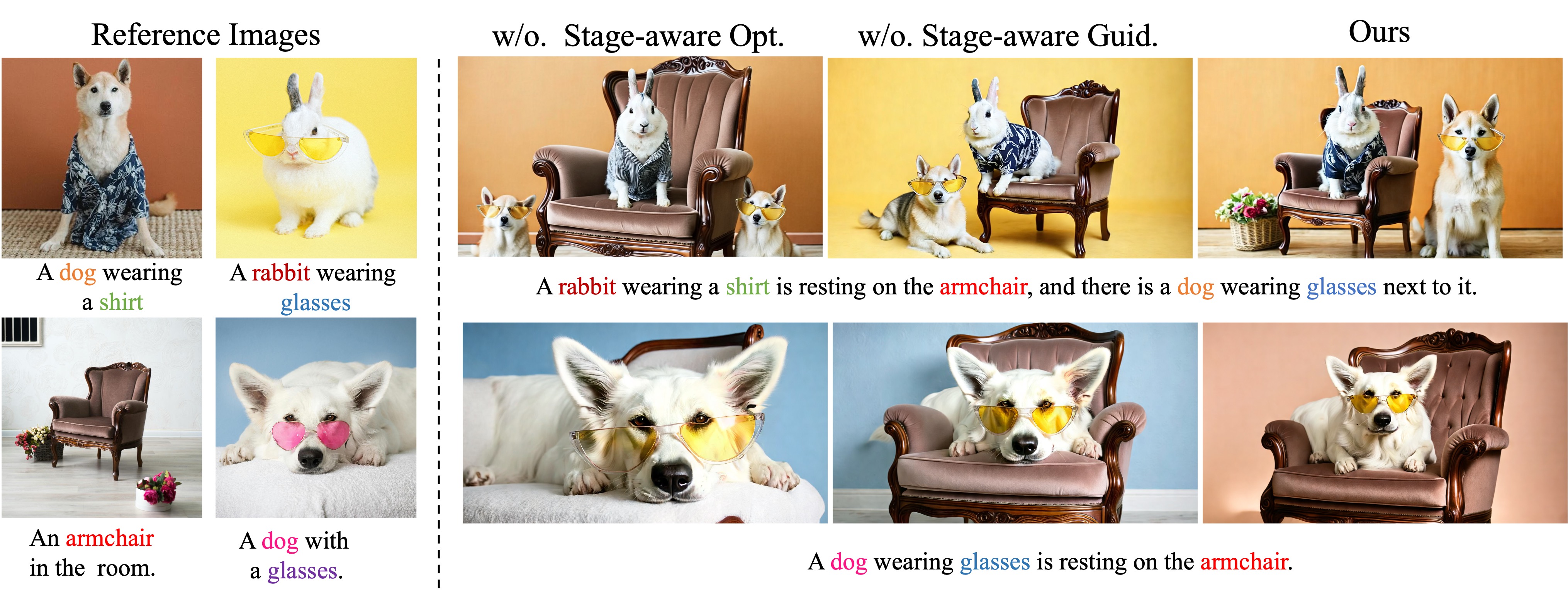}
    \vspace{-10pt}	
    \caption{\textbf{Qualitative Ablation of Stage-Aware Components.} We compare our full model with variants lacking key components. (a) w/o Stage-aware Opt.: Naive training fails to learn a high-fidelity concept. (b) w/o Stage-aware Guid.: Applying guidance uniformly across all stages causes structural artifacts and concept bleeding. (c) Ours: The synergy of both strategies yields the best compositional quality.}
    \label{fig:ablation_components}
    \vspace{-5pt}	
\end{figure*}
\vspace{-5pt}	
\begin{figure}[ht!]
    \centering
    \includegraphics[width=0.9\linewidth]{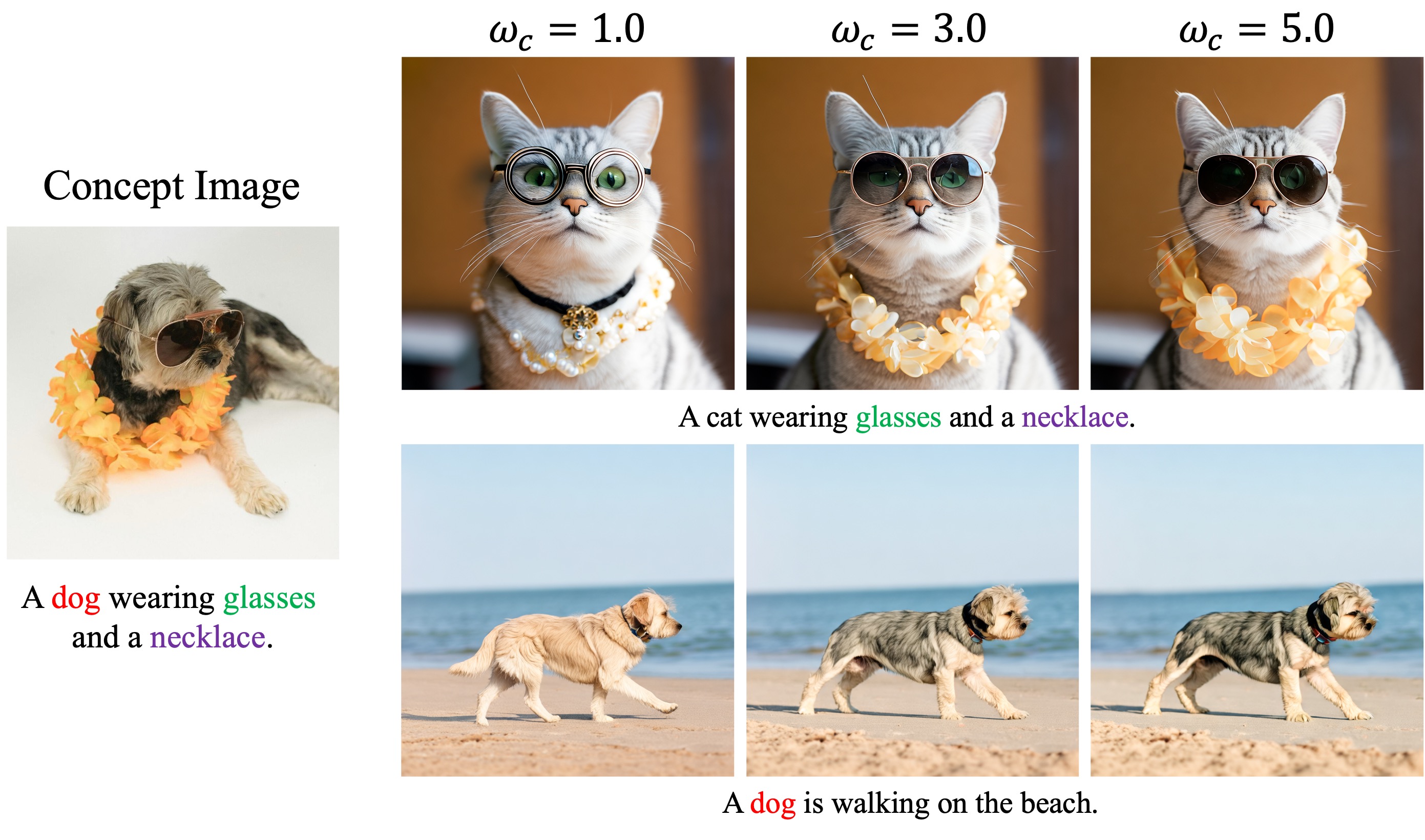}
    \vspace{-10pt}	
    \caption{\textbf{Analysis of the Concept Guidance Scale $w_c$.} As $w_c$ increases, the visual attributes of the generated concepts align more closely with the reference image. This demonstrates that $w_c$ provides fine-grained control over the strength of the personalization.}
    \label{fig:ablation_scale}
    \vspace{-20pt}	
\end{figure}

\subsection{User Study}
We conducted a user study with 10 participants using a 5-point Likert scale (1–5) to evaluate \textbf{Concept Consistency}, \textbf{Text Adherence}, and \textbf{Aesthetic Quality}. As shown in Table \ref{tab:user_study_results}, ConceptWeaver achieves the highest average scores across all metrics. Notably, it scores \textbf{4.06} in Concept Consistency, significantly outperforming MS Diffusion (3.75) and Dreambooth (3.02). Similar improvements are observed in Text Adherence (\textbf{3.78}) and Aesthetic Quality (\textbf{4.12}), confirming ConceptWeaver's superior ability to balance concept fidelity with prompt controllability and visual appeal.

\begin{table*}[ht]
    \vspace{-15pt}
    \centering
    \caption{User Study Results for Concept Generation (Values represent average scores on a 1-5 scale).}
    \label{tab:user_study_results}
    \begin{tabular}{lccc}
        \toprule
        Metric & \textbf{ConceptWeaver} & MS Diffusion & Dreambooth \\
        \midrule
        Concept Consistency & \textbf{4.06} & 3.75 & 3.02 \\
        Text Adherence      & \textbf{3.78} & 3.14 & 2.88 \\
        Aesthetic Quality   & \textbf{4.12} & 3.75 & 3.05 \\
        \bottomrule
    \end{tabular}
    \vspace{-15pt}
\end{table*}

\subsection{Ablation Study}
\label{sec:ablation}

To validate the effectiveness of our proposed components and understand their contributions, we conduct a series of ablation studies. We first analyze the necessity of our stage-aware mechanisms and then investigate the effect of the concept guidance scale $w_{c,i}$.

\paragraph{Effect of Stage-Aware Components.}
We compare our full model against two variants: (1) \textbf{w/o Stage-aware Opt.}, where the concept offset is trained uniformly over all timesteps; and (2) \textbf{w/o Stage-aware Guid.}, where the guidance is applied uniformly across all three stages. As shown in Figure~\ref{fig:ablation_components}, our full model achieves the best visual quality. Without stage-aware optimization, the learned concept lacks fidelity. Without stage-aware guidance, noticeable artifacts and concept bleeding appear, as the guidance interferes with structure formation and detail refinement. This confirms that both stage-aware components are integral and work synergistically. More visualization examples are shown in the Appendix \textbf{C}. 

\vspace{-10pt}

\paragraph{Analysis of Concept Guidance Scale.}
The concept guidance scale $w_{c,i}$ in Eq.~\ref{eq:cwg_multi} balances the influence of the learned concept against the base text prompt. We analyze its effect in Figure~\ref{fig:ablation_scale}. When $w_c$ is low (e.g., $w_c=1.0$), the model generates a generic version of the concept based on the text prompt. As $w_c$ increases, the visual attributes of the generated concepts (e.g., the specific style of the *glasses* and *necklace*, or the identity of the *dog*) align more closely with the reference image. This demonstrates that $w_c$ provides a fine-grained, user-controllable knob for modulating the strength of personalization and achieving the desired visual fidelity. More examples are shown in the Appendix \textbf{D}. 

\begin{figure}[ht!]
    \centering
    \includegraphics[width=1.0\linewidth]{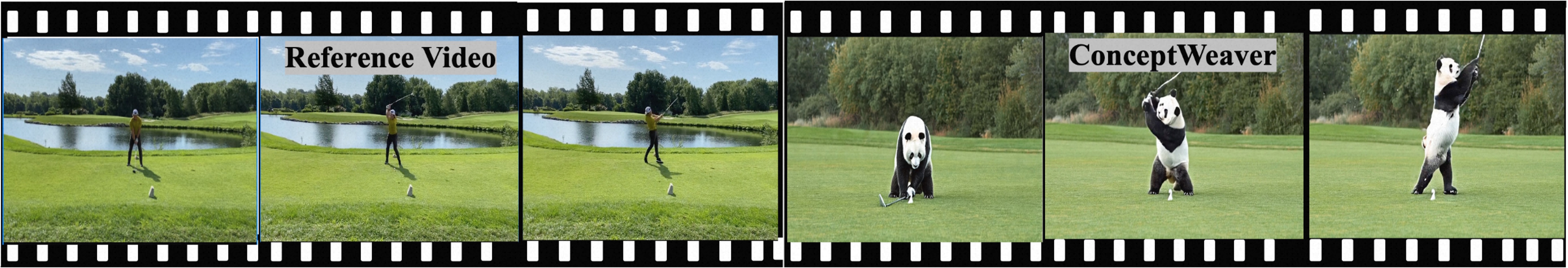}
    \vspace{-20pt}	
    \caption{\textbf{Motion Concept Transfer with ConceptWeaver.} We learn a motion concept from a single reference video (up) and apply it to new subjects defined by text prompts (down). The generated videos demonstrate high-fidelity motion replication while accurately synthesizing the new subject's appearance, showcasing effective disentanglement of motion and content.}
    \label{fig:motiontransfer}
    \vspace{-15pt}	
\end{figure}

\begin{figure}[ht!]
    \centering
    \includegraphics[width=1.0\linewidth]{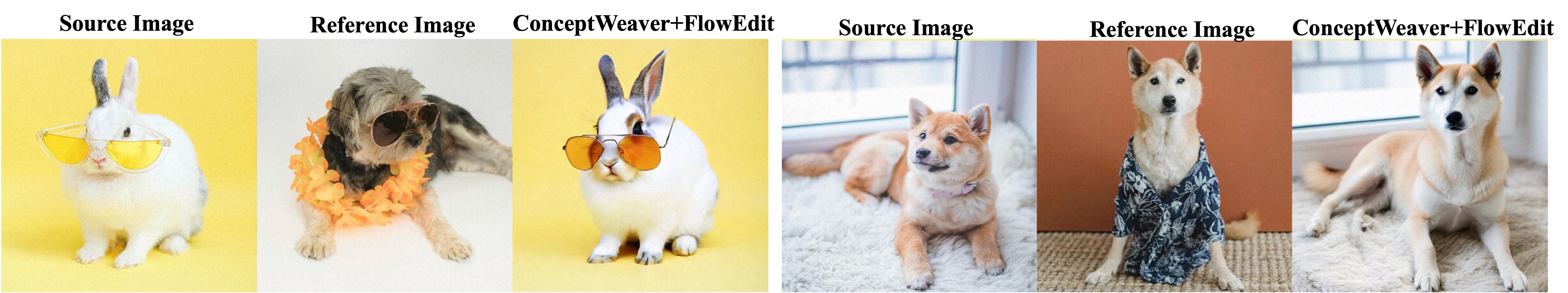}
    \vspace{-15pt}	
    \caption{\textbf{Image Editing with ConceptWeaver and FlowEdit.} By integrating with FlowEdit, we apply a visual concept learned from a reference image to a source image. The resulting edited image preserves the source's structure while faithfully incorporating the specific appearance of the learned concept.}
    \label{fig:flowedit}
    \vspace{-20pt}	
\end{figure}

\subsection{Applications}

\paragraph{Motion Concept Transfer.} Our framework achieves one-shot motion transfer by treating motion as a structural concept. As our probing (Sec.~\ref{sec:probing}) shows its influence peaks in the Blueprint Stage, we apply our stage-aware optimization and guidance for motion exclusively in this early stage. This effectively decouples motion from appearance, enabling the high-fidelity transfer of dynamics from a reference video to new subjects, as illustrated in Figure~\ref{fig:motiontransfer}. More visualization examples are shown in the Appendix \textbf{E}. 

\vspace{-10pt}

\paragraph{Customized Image Editing.}
ConceptWeaver can be extended to real image editing by integrating with the FlowEdit \cite{kulikov2025flowedit}. We first learn a concept offset $\Delta \boldsymbol{c}_{\mathrm{ref}}$ from a reference image. This offset is then used to create a customized target velocity, $\boldsymbol{v}_{\mathrm{target}} = \boldsymbol{v}_\theta(\boldsymbol{x}_t, t, \boldsymbol{c}_{\mathrm{edit}} + \Delta \boldsymbol{c}_{\mathrm{ref}})$, which guides the FlowEdit process. This synergy allows FlowEdit to preserve the source image's structure while our stage-aware guidance injects the high-fidelity appearance of the learned concept. As shown in Figure~\ref{fig:flowedit}, this enables customized, one-shot editing of real images. More visualization examples are shown in the Appendix \textbf{F}.

%% file: sections/5_conclusion.tex
\vspace{-10pt}
\section{Conclusion}
\vspace{-5pt}
In this work, we addressed the challenge of one-shot concept decoupling in flow models. We introduced a novel differential probing technique that revealed a universal three-stage generative framework---\textbf{Blueprint}, \textbf{Instantiation}, and \textbf{Refinement}---which governs how concepts are formed over time. This key insight demonstrated that concepts naturally disentangle during a pivotal mid-stage, providing a principled and highly effective window for manipulation. Guided by this discovery, we proposed ConceptWeaver, a framework that operationalizes our findings. It learns concept-specific offsets via a stage-aware optimization and applies them with a CWG mechanism. By aligning closely with the model's intrinsic dynamics rather than forcing external constraints, ConceptWeaver achieves high-fidelity compositional control from a single reference image. Our work highlights a shift towards principled, analysis-driven model control. While we manually categorize concepts, a promising future direction is to learn these temporal profiles automatically. We believe that understanding and leveraging the inherent structure of generative models is the key to unlocking the next frontier of precise and creative \mbox{content synthesis}.

%% file: sections/X_supp.tex
\clearpage
\appendix
\begin{center}
    {\Large \bfseries Supplementary Material}
\end{center}

\setcounter{section}{0}
\renewcommand{\thesection}{\Alph{section}}

\section{Baselines}
\label{sec:Baselines}
For the replication of TokenVerse, since there is no modulation process for the text space in DiT-like architectures of Wan2.2-5B, we introduce an MLP to generate the corresponding shift and scale to incorporate this modulation. Following the original paper's methodology, for the first 800 training steps, we sample 92\% from the 800-1000 timestep range and 8\% from the 0-800 timestep range. For the subsequent 1200 training steps, this sampling ratio is reversed. For DreamBooth, we employ the LoRA training method and utilize 10\% prior loss.

For ConceptWeaver, we set the total denoising steps to 30. These steps are divided into three stages: the first 10 steps for the \textbf{Blueprint Stage}, the middle 10 steps for the \textbf{Instantiation Stage}, and the final 10 steps for the \textbf{Refinement Stage}. For content concepts, we set the CWG guidance weights for these three stages to 0, 8, and 3, respectively. For structural concepts, the CWG guidance weights are set to 5, 1, and 0. All of our experiments were performed on a single \texttt{H20} GPU.

%

\section{More Results}
\subsection{Single-Concept Generation}
\begin{figure}[t]
    \centering
    \includegraphics[width=0.90\linewidth]{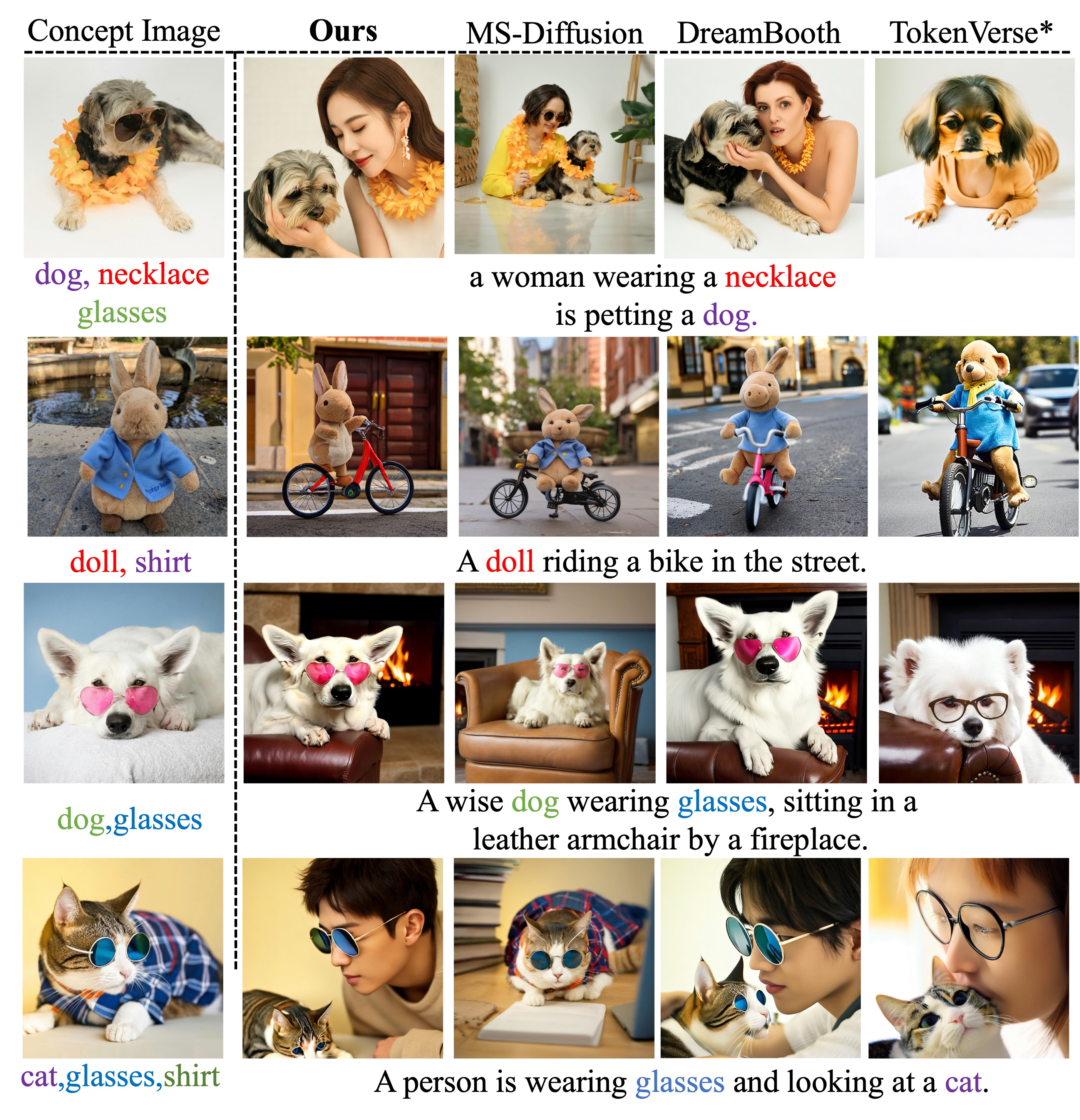}
    \vspace{-10pt}	
    \caption{\textbf{Qualitative results on single-concept generation.} Each row showcases results for a specific learned concept. The first column displays the reference image used to learn the concept. Subsequent columns display generated images from different methods based on a new text prompt. }
    \label{fig:single_comp}
    \vspace{-15pt}	
\end{figure}

We present a qualitative comparison for single-concept generation in Figure~\ref{fig:single_comp}, which supplements the quantitative findings in the main paper's Table 1. In these side-by-side comparisons with leading methods such as MS-Diffusion, DreamBooth, and TokenVerse*, our ConceptWeaver demonstrates a superior ability to disentangle and replicate concepts with high fidelity. While other methods often struggle with concept bleeding (e.g., DreamBooth often suffers from context overfitting, causing elements from the reference image's background to 'bleed' into the new generation) or fail to capture specific attributes (e.g., TokenVerse* generating a different style of glasses), our approach consistently preserves the nuanced visual details of the reference concept across diverse prompts and contexts, validating its effectiveness in high-fidelity personalization.

\subsection{Multi-Concepts Generation}
As an extension of the multi-concept composition results shown in the main paper (Figure 5), we provide more extensive qualitative comparisons in Figure~\ref{fig:sup_multi}. These results highlight ConceptWeaver's superior performance in disentangling and composing multiple concepts from different reference images. Compared to baseline methods, our approach significantly reduces concept bleeding and overfitting, resulting in more coherent and plausible compositions that accurately reflect the target prompt.

\section{Ablation Study of Stage-Aware Components}
To validate the critical role of our stage-aware mechanisms, we conduct a detailed ablation study, with additional qualitative results presented in Figure~\ref{fig:sup_stage}. These examples, which supplement Figure~7 from the main paper, reveal a crucial aspect beyond mere visual fidelity: \textbf{pose disentanglement}. The comparisons clearly show that removing either stage-aware component leads to significant degradation. Specifically, the model trained without \textbf{Stage-aware Optimization} not only fails to capture the detailed appearance of the concept but also tends to rigidly replicate the pose from the reference image. For instance, in the last row, the dog remains in a static lying-down pose, unable to adopt the dynamic running pose suggested by the context. Similarly, applying guidance uniformly (\textbf{w/o Stage-aware Guid.}) results in concept interference and, more importantly, strong pose overfitting. The dog in the second row, for example, is persistently generated in a seated position, mirroring the reference image, regardless of the new beach context. In stark contrast, our \textbf{full model (Ours)} successfully disentangles the concept's identity and attributes from its original pose. It can generate the learned concepts in novel, context-appropriate poses—such as the dog lying down on the beach or running in the snow. This demonstrates that the synergy of stage-aware optimization and guidance is essential not just for high-fidelity rendering, but for achieving true compositional generalization where concepts can be freely manipulated in new spatial arrangements.

\section{Ablation Study of CWG}

In this section, we provide additional visualizations to supplement the analysis of our ConceptWeaver Guidance (CWG) mechanism, first presented in Figure~8 of the main paper. Figure~\ref{fig:sup_cwg} showcases how varying the guidance scales for different learned concepts provides fine-grained control over the final synthesis. 

To specifically isolate the effect of the pivotal \textbf{Instantiation Stage}, we fix the guidance scales for the \textbf{Blueprint Stage} at a low value (to allow for structural flexibility) and for the \textbf{Refinement Stage} at a moderate value (to ensure detail coherence). We then incrementally increase the guidance scales ($g_1, g_2$) for the two learned content concepts, which are applied only during their peak-influence \textbf{Instantiation Stage}.

The results demonstrate a clear and intuitive progression. 
\begin{itemize}
    \item \textbf{At low guidance values} ($g_1=0, g_2=0$), the model generates generic concepts primarily driven by the text prompt, failing to incorporate the specific visual attributes from the reference images. For example, the woman wears generic glasses, and the dog wears a generic hat.
    \item \textbf{As the guidance scales increase} (from $g_1, g_2=3$ to $8$), the visual characteristics of the output progressively conform to the reference concepts. In the top row, the generic glasses transform into the specific yellow-tinted glasses from the reference rabbit. In the second row, the reference hat materializes on the dog's head. Similarly, the specific identity of the teddy bear \textit{doll} and the unique frame of the \textit{bicycle} are gradually and faithfully rendered.
    \item \textbf{Optimal Fidelity}: For most cases, optimal visual fidelity is achieved at higher guidance values (e.g., $g_1=8, g_2=8$), where the generated concepts are nearly identical to the reference ones, yet are seamlessly integrated into a completely new scene and composition.
\end{itemize}
This controlled, gradual refinement validates two key aspects of our framework: first, that the \textbf{Instantiation Stage} is indeed the optimal window for injecting content identity; and second, that CWG serves as an effective and intuitive control knob for users to modulate the strength of personalization and achieve the desired balance between prompt-adherence and concept fidelity.

\begin{figure*}[t]
    \centering
    \includegraphics[width=0.90\linewidth]{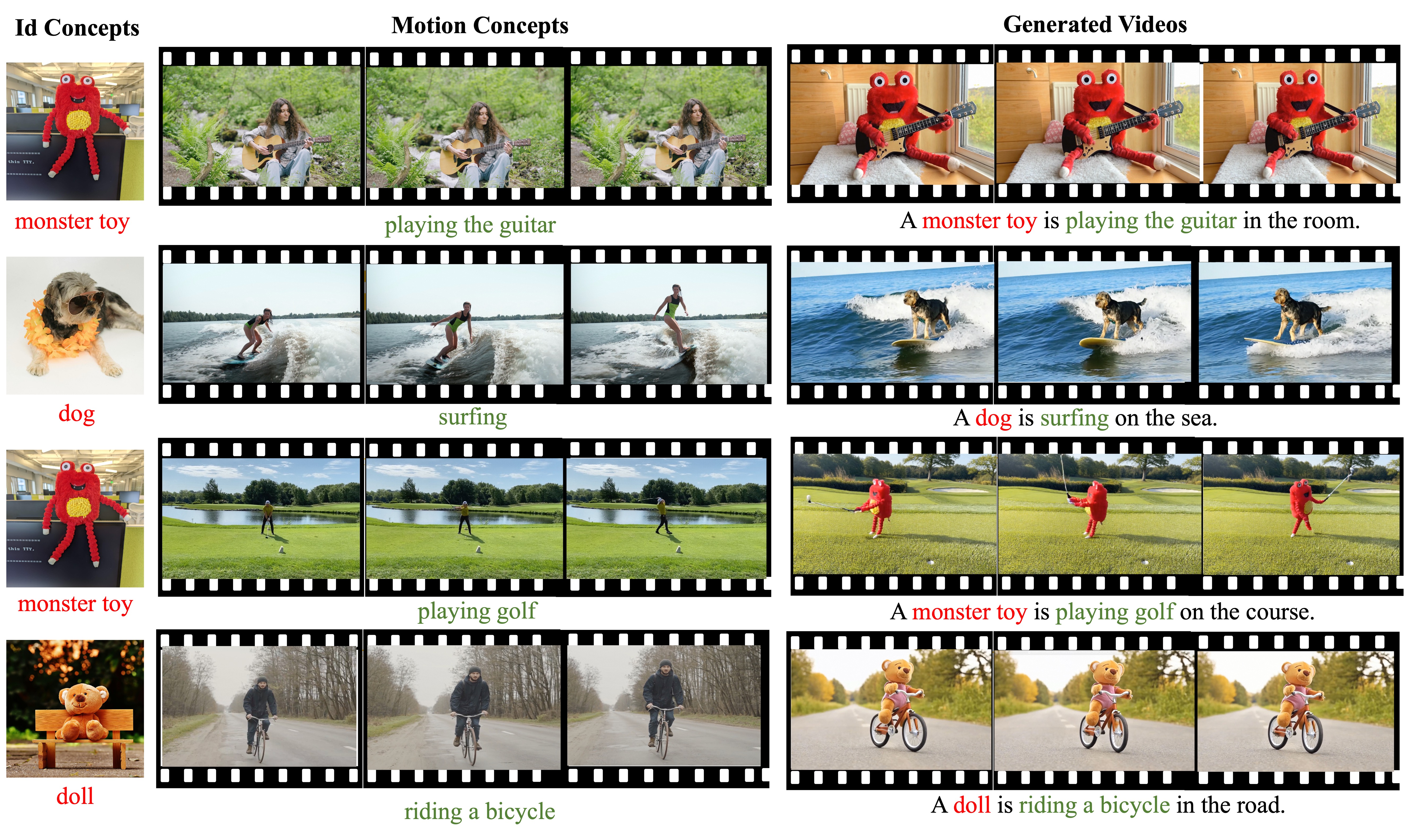}
    \vspace{-10pt}
    \caption{\textbf{Qualitative results on motion transfer.} Each row demonstrates the ability of ConceptWeaver to compose a subject's identity, learned from a static \textbf{Id Concept} image, with complex dynamics learned from a \textbf{Motion Concepts} video. The generated videos successfully apply the reference motion (e.g., "surfing", "playing golf") to a completely different subject (e.g., a "dog", a "monster toy") while maintaining high subject fidelity and generating a plausible new scene. This highlights our method's strong capability for disentangling and recombining structural (motion) and content (identity) concepts.}
    \label{fig:motion2}
    \vspace{-20pt}	
\end{figure*}
\section{Motion Concept Transfer}

Our framework's ability to treat motion as a structural concept enables high-fidelity, one-shot motion transfer. By optimizing and applying the motion concept offset exclusively during the Blueprint Stage, we effectively disentangle motion from appearance. We present additional examples of motion concept transfer in Figure~\ref{fig:motion2}. These results, which supplement Figure 9 in the main paper, show that complex dynamics learned from a single reference video can be successfully applied to diverse new subjects, while accurately synthesizing their appearances according to the text prompt.

\section{Customized Image Editing}

By integrating ConceptWeaver with FlowEdit, our framework can be extended from generative synthesis to powerful real-image editing tasks, enabling one-shot appearance transfer. As shown in Figure~\ref{fig:sup_flowedit} (and supplemented by Figure~7 in the main paper), the process involves first learning a concept offset $\Delta \boldsymbol{c}_{\text{ref}}$ from a \textbf{Reference Image}. This offset, which encapsulates the disentangled visual essence of the concept, is then used to define a customized target velocity, $\boldsymbol{v}_{\text{target}} = \boldsymbol{v}_{\theta}(\boldsymbol{x}_t, t, \boldsymbol{c}_{\text{edit}} + \Delta \boldsymbol{c}_{\text{ref}})$. This customized velocity guides the FlowEdit process, effectively "painting" the learned appearance onto a \textbf{Source Image}.

The results demonstrate a remarkable synergy between the two methods. FlowEdit excels at preserving the original image's structure, layout, and composition. Simultaneously, our stage-aware ConceptWeaver Guidance precisely injects the high-fidelity appearance of the learned concept. This allows for targeted edits, such as changing a modern bicycle to a vintage one while keeping it on the same street, or transforming a puppy's breed while it remains sitting on the same patch of grass.

Crucially, this successful integration is underpinned by our core discovery of the three-stage generative process. The FlowEdit inversion and editing process is typically configured to operate within the early-to-mid diffusion timesteps (e.g., steps 15 to 25 out of 30 in our setup), as this window is optimal for altering high-level semantics without disrupting fine details. This aligns perfectly with our finding that the \textbf{Instantiation Stage}—the peak influence window for content concepts—occurs in this exact mid-range. By injecting our learned concept offset precisely during this shared, optimal window, ConceptWeaver ensures that the concept's appearance is applied with maximum impact and minimal interference, leading to the seamless and high-quality edits shown.

\begin{figure}[h]
    \centering
    \vspace{-15pt}
    \includegraphics[width=0.90\linewidth]{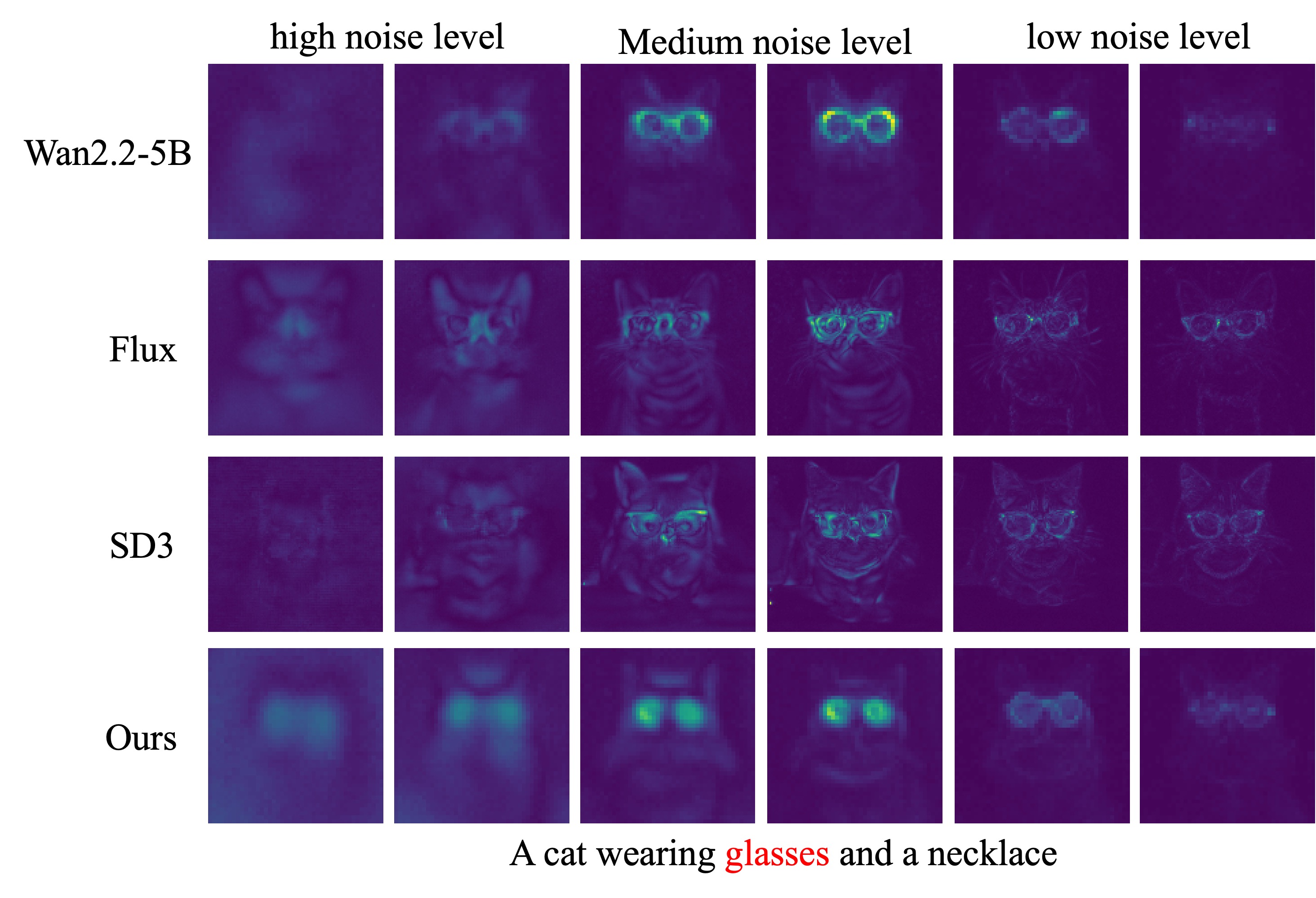}	
    \vspace{-10pt}
    \caption{\textbf{Generalization of the Three-Stage Framework Across Different Flow Models.} Our differential probing of the "glasses" concept reveals a universal three-stage generative process across diverse flow models (\textbf{Wan2.2-5B}, \textbf{Flux}, \textbf{SD3}, and \textbf{Ours}). All models consistently show a weak signal during the early \textbf{Blueprint Stage}, a sharp peak of influence during the mid-\textbf{Instantiation Stage}, and signal decay during the final \textbf{Refinement Stage}. This confirms that our discovered framework is an intrinsic property of flow-based generation, not a model-specific artifact.}
    \label{fig:vs-glasses}	
    \vspace{-20pt}
\end{figure}
\section{Generalization Study}

A cornerstone of our work is the discovery of the three-stage generative framework. To validate that this is a fundamental property of flow-based models rather than an artifact of a specific architecture, we extended our differential probing analysis (Sec. 4.1) to a diverse set of prominent models. As shown in Figure~\ref{fig:vs-glasses}, we applied our probing technique to visualize the emergence of the concept "glasses" in \textbf{Wan2.2-5B}, \textbf{Flux}, and \textbf{Stable Diffusion 3 (SD3)}. These models represent a wide spectrum of design philosophies, from large-scale transformers to different conditioning mechanisms.

The results reveal a remarkably consistent, non-monotonic lifecycle for the concept's influence across all tested architectures. For each model, we observe the same distinct phases:
\begin{itemize}
    \item \textbf{Blueprint Stage (high noise):} In the early steps of generation, the velocity shift corresponding to "glasses" is weak and spatially diffuse. The models appear to be establishing the coarse structure of the scene (e.g., the general head shape of a cat) before committing to specific details.
    \item \textbf{Instantiation Stage (medium noise):} During the pivotal mid-stage of generation, the signal for "glasses" experiences a sharp peak in both intensity and spatial localization. The shape of the glasses becomes clearly defined and is accurately placed on the cat's face. This phase represents the critical window where the concept is fully materialized and naturally disentangled, making it ideal for manipulation.
    \item \textbf{Refinement Stage (low noise):} In the final stages, the concept-specific signal diminishes significantly. The models' focus shifts from concept formation to harmonizing fine-grained textures, adjusting lighting, and ensuring global coherence of the entire image.
\end{itemize}
This striking cross-model consistency provides strong evidence that the three-stage framework is an intrinsic and generalizable characteristic of conditional flow-based generation. It suggests that our analysis-driven approach to concept control, which leverages this temporal structure, has the potential for broad applicability across the rapidly evolving ecosystem of flow models.

\section{Limitations}
While our ConceptWeaver method demonstrates significant advancements in personalized image generation, it is inherently subject to certain limitations primarily stemming from the capabilities of the underlying generative model. Generally, the stronger the base model's generative capacity, the better the quality of the synthesized concepts. This dependency implies that our method's performance will scale with the advancements in generation models.

\begin{figure}[h]
    \centering
    \vspace{-15pt}
    \includegraphics[width=0.90\linewidth]{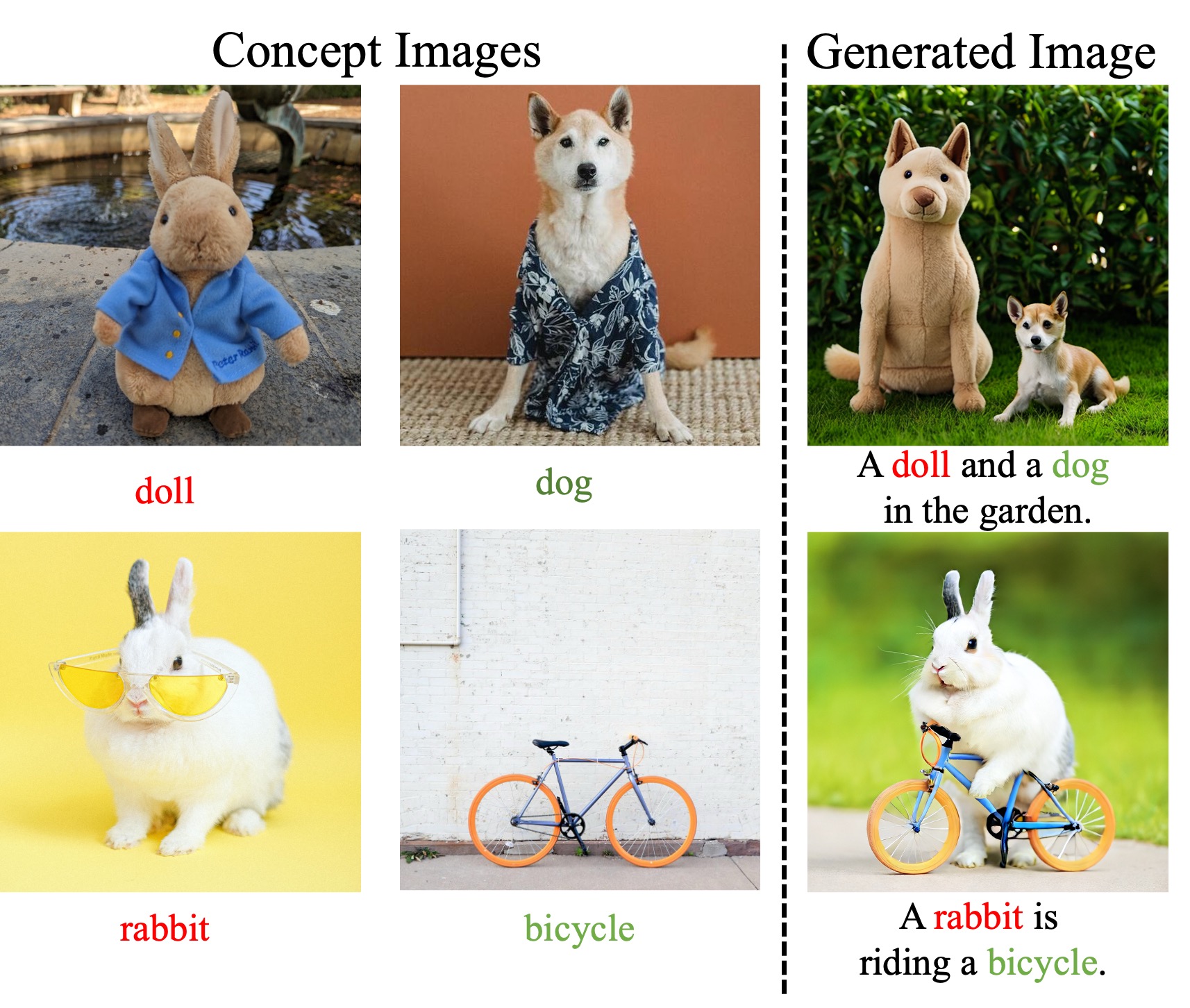}	
    \vspace{-15pt}
    \caption{\textbf{Limitations.} \textbf{The first row} illustrates \emph{Concept Semantic Confusion}, where similar concepts (e.g., dog and doll) lead to undesirable infiltration of one concept into another during generation. \textbf{The second row} demonstrates scenarios where the desired generation task or concept complexity \emph{Exceeds Underlying Model Capabilities}, exemplified by a rabbit attempting human-like riding motions.}
    \label{fig:limitation}	
    \vspace{-15pt}
\end{figure}
As shown in Figure \ref{fig:limitation}, our method encounters limitations in two main scenarios. First, \textbf{concept semantic confusion} (first row) arises when confronted with very similar concepts, leading to less distinct representations. Second, the model struggles with tasks that \textbf{exceed its foundational capabilities} (second row), resulting in unsatisfactory or erroneous outputs. Future work will focus on addressing these issues through stronger concept disentanglement and more powerful base generative models.

\begin{figure*}[t]
  \centering
  \includegraphics[width=1.0\linewidth]{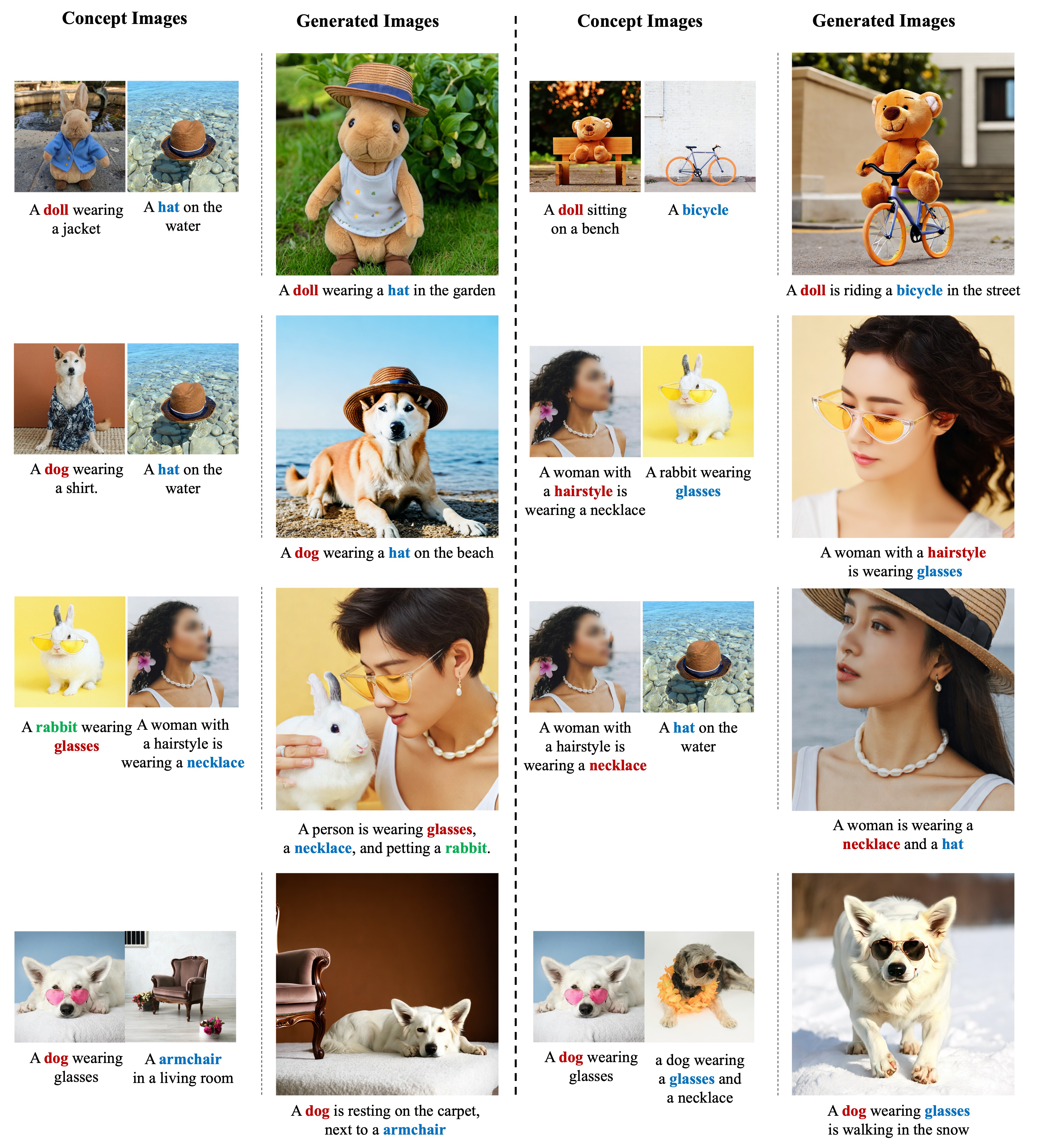}
  \caption{\textbf{Qualitative results on multi-concept composition generation.} Our method successfully disentangles and composes multiple concepts from different source images into novel scenes. Each panel shows concepts (e.g., a specific \textit{doll}, a \textit{hat}, \textit{glasses}, a \textit{hairstyle}) learned from one or more reference images (Concept Images) and then synthesized according to a new prompt (Generated Images).}
  \label{fig:sup_multi}
  \vspace{-10pt}
\end{figure*}

\begin{figure}[h]
  \centering
  \includegraphics[width=1.0\linewidth]{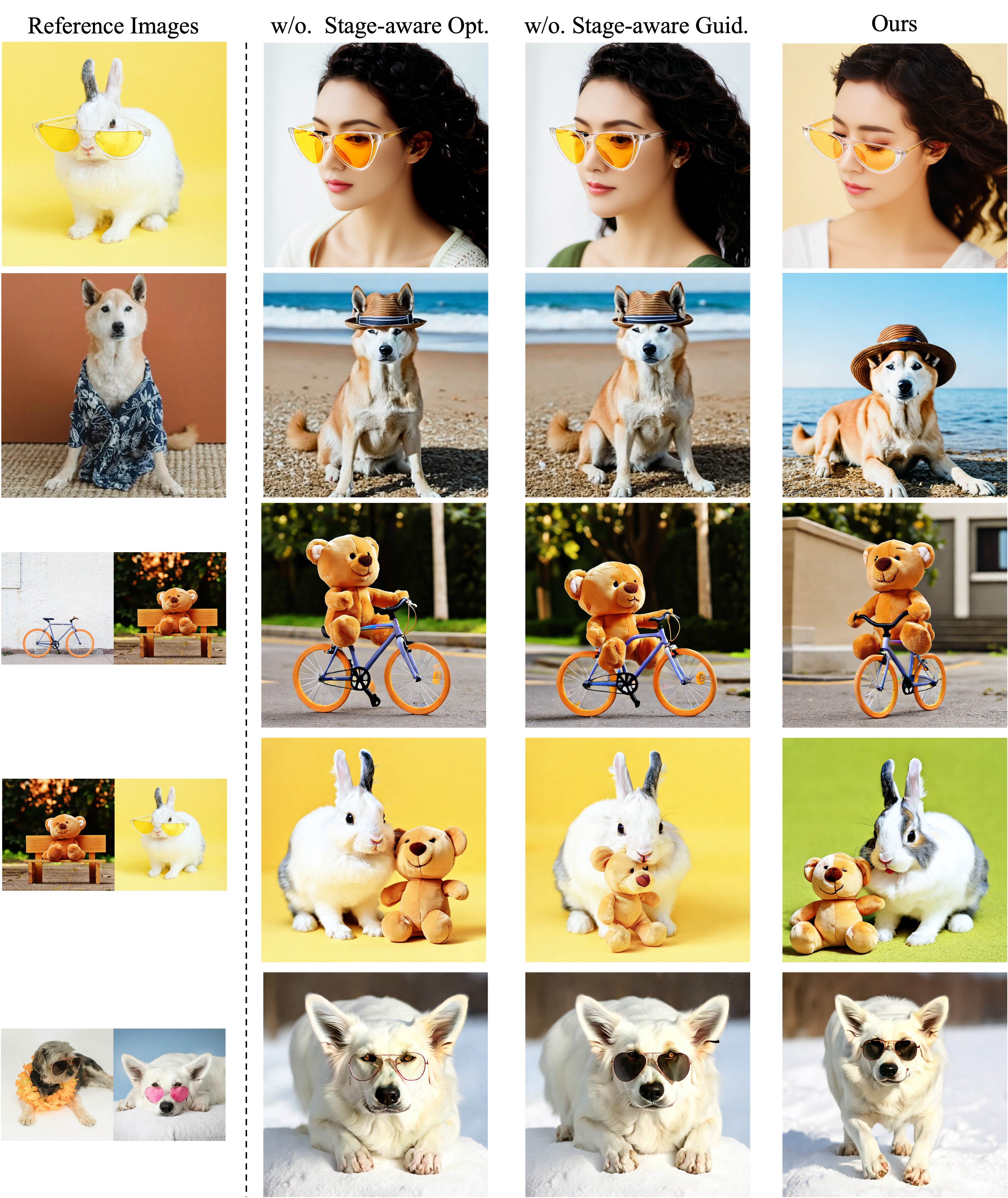}
  \caption{\textbf{Qualitative results on Stage-Aware Components.} This comparison highlights how our stage-aware mechanisms are crucial for both fidelity and pose generalization. Both ablated versions, \textbf{w/o Stage-aware Opt.} and \textbf{w/o Stage-aware Guid.}, exhibit strong pose overfitting, rigidly replicating the posture of the subject from the reference image (e.g., the seated dog in the second row, the lying-down dog in the last row). In contrast, our \textbf{full model (Ours)} successfully disentangles the concept from its original pose, allowing it to be rendered in new, contextually appropriate poses, such as running or lying down. This demonstrates that our method achieves a higher level of compositional control.}
  \label{fig:sup_stage}
  \vspace{-10pt}
\end{figure}

\begin{figure}[t]
  \centering
  \includegraphics[width=1.0\linewidth]{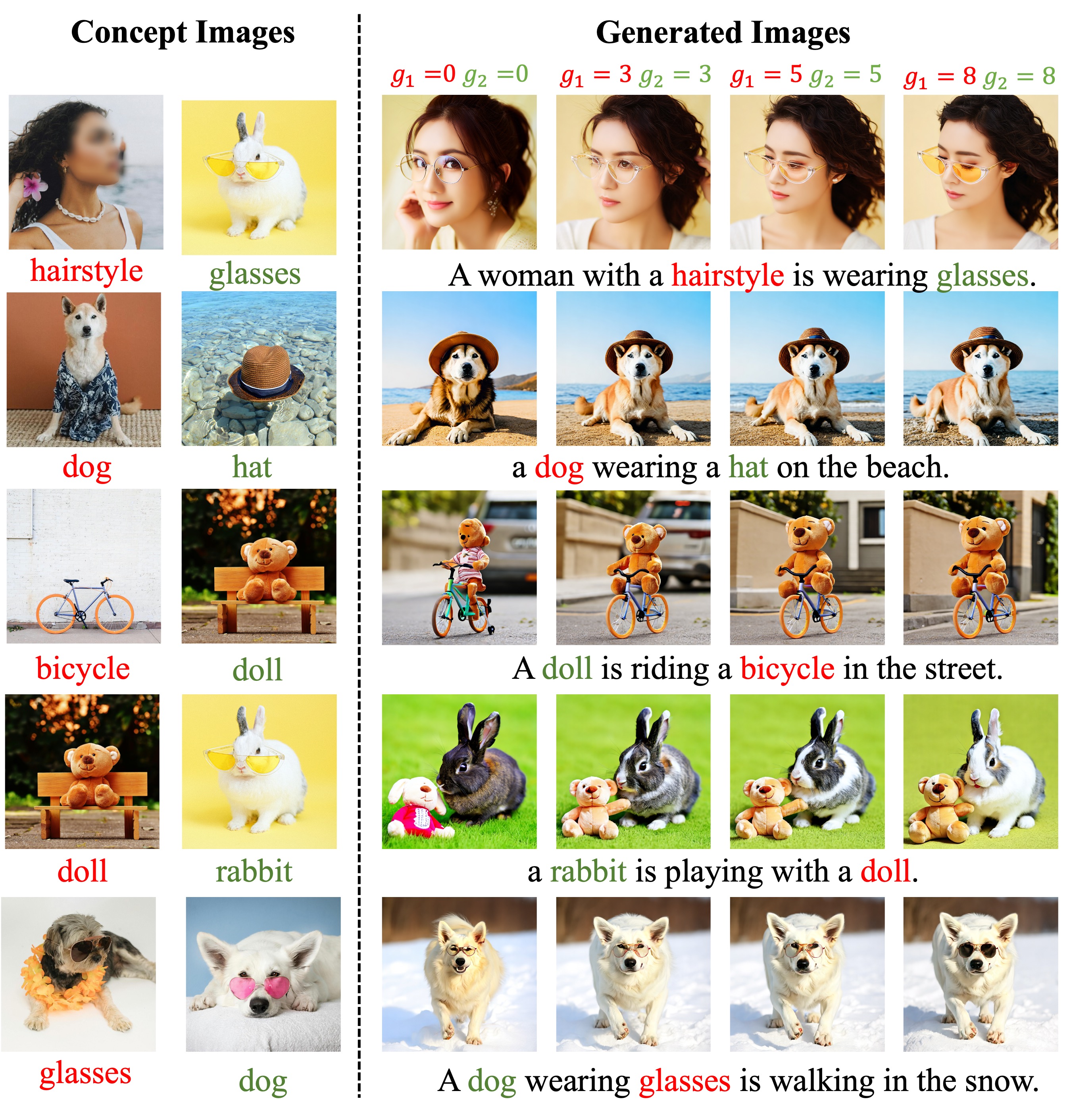}
  \caption{\textbf{Qualitative results on ConceptWeaver Guidance}. We fix the CWG guidance for the first stage at $g_1=0, g_2=0$ and for the third stage at $g_1=3, g_2=3$, to qualitatively analyze our CWG. From the results, it can be observed that the optimal effect is achieved when the CWG value in the second stage is 8, and there is a clear phenomenon of concept gradual refinement as CWG increases.}
  \label{fig:sup_cwg}
  \vspace{-10pt}
\end{figure}

\begin{figure}[t]
  \centering
  \includegraphics[width=0.90\linewidth]{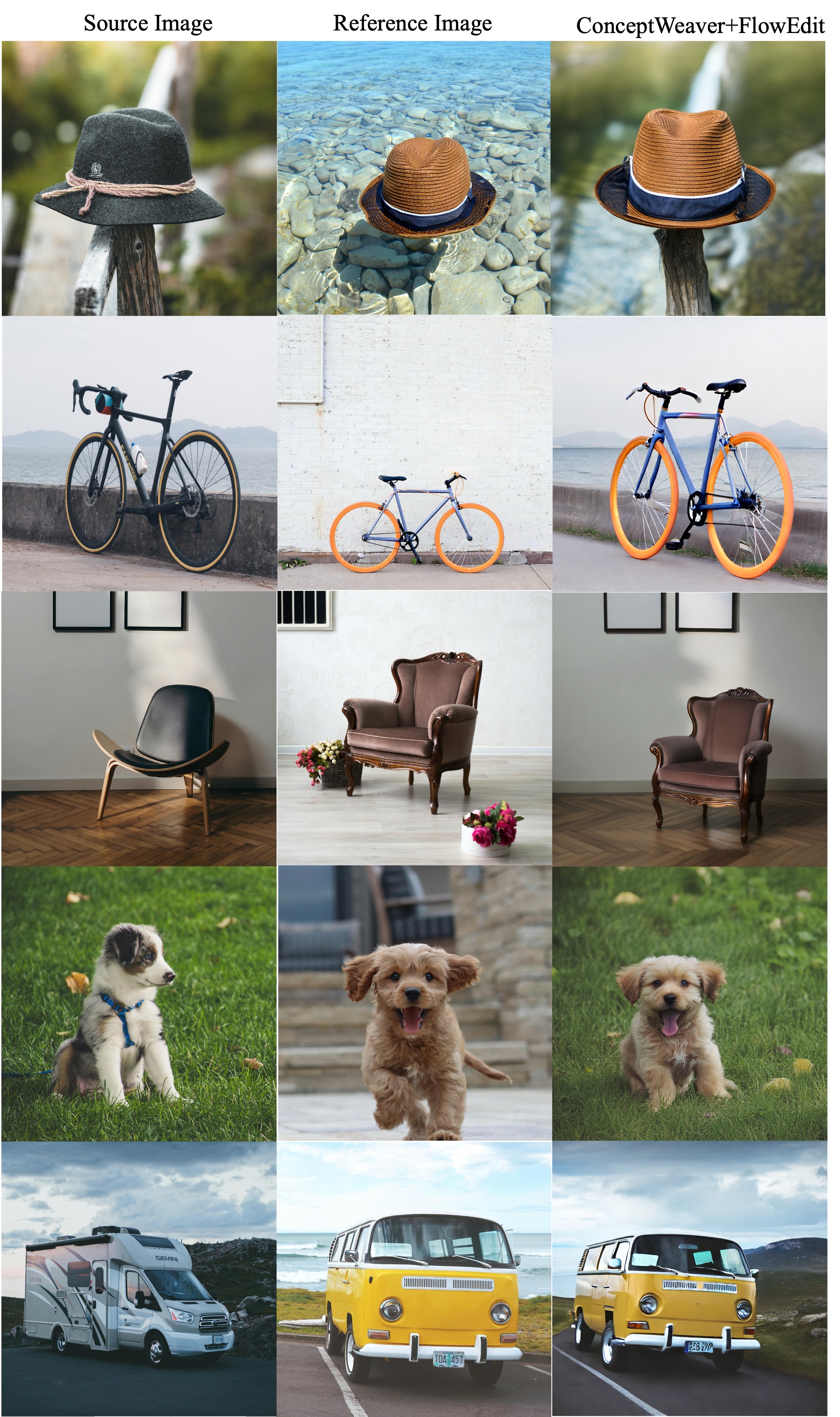}
  \vspace{-10pt}
  \caption{\textbf{Image Editing with ConceptWeaver and FlowEdit.} By integrating with FlowEdit, ConceptWeaver enables precise appearance swapping. Our method first disentangles a target concept's appearance from a Reference Image—like the texture of a straw hat or the look of a VW van.}
  \label{fig:sup_flowedit}
\end{figure}